%% file: 00_main.tex
\documentclass{article} % For LaTeX2e
\usepackage{iclr2026_conference,times}
\iclrfinalcopy

\input{math_commands.tex}

\usepackage{hyperref}
\usepackage{url}
\usepackage{graphicx}
\usepackage{booktabs}
\usepackage{xcolor}
\usepackage{adjustbox}
\usepackage{rotating}
\usepackage{amsmath}
\usepackage{amsthm}
\usepackage{bm} 
\usepackage{siunitx}
\usepackage{caption}
\usepackage{tikz}
\usepackage{booktabs}
\usepackage{subcaption}
\usepackage{tabularx}
\usepackage{tablefootnote}
\usepackage{wrapfig}

\usetikzlibrary{positioning,fit,backgrounds,intersections,tikzmark}
\usepackage{xcolor,colortbl}

\usepackage[table]{xcolor}
\definecolor{lightgreen}{HTML}{d0f0d0} 

\usepackage{hyperref}
\hypersetup{
  colorlinks   = true,
  linkcolor    = red!70!black,
  citecolor    = teal!70!black,
  urlcolor     = red!70!black,
  filecolor    = gray
}

\newtheorem{theorem}{Theorem}

\newtheorem{proposition}[theorem]{Proposition}

\definecolor{colorIMD}{RGB}{147, 112, 219}
\definecolor{colorOoMD}{RGB}{223, 83, 83}

\definecolor{lightyellow}{RGB}{255, 255, 204} 

\makeatletter
\def\@maketitle{\vbox{\hsize\textwidth
  {\LARGE\sc \@title\par}
  \vskip 0.15in
  \centering
  \begin{tabular}[t]{c}\bf\rule{\z@}{24pt}\ignorespaces
    \@author
  \end{tabular}%
  \vskip 0.3in
}}
\makeatother

\title{\centering Meta-Statistical Learning: Supervised Learning of Statistical Estimators}

\author{%
  \begin{tabular}{c}
    \begin{tabular}{ccc}
      Maxime Peyrard\textsuperscript{$\diamondsuit$} &
        Kyunghyun Cho\textsuperscript{$\clubsuit$}\\
    \end{tabular} \\[6pt]
    \normalfont
    \textsuperscript{$\diamondsuit$}Universit\'e Grenoble Alpes, CNRS, Grenoble INP, LIG \\ 
    \normalfont
    \textsuperscript{$\clubsuit$}New York University, Prescient Design, Genentech \\
    \normalfont
    \small
    \texttt{peyrardm@univ-grenoble-alpes.fr}\\
    \normalfont
    \small
    \texttt{kyunghyun.cho@nyu.edu}\\
  \end{tabular}%
}

\begin{document}

%% Comment colors 
\newcommand{\Megan}[1]{\textcolor{blue}{[Megan: #1]}}
\newcommand{\MaximeM}[1]{\textcolor{green}{[Maxime M: #1]}}
\newcommand{\German}[1]{\textcolor{orange}{[German: #1]}}
\newcommand{\MaximeP}[1]{\textcolor{purple}{[Maxime P: #1]}}
\newcommand{\Kyunghyun}[1]{\textcolor{red}{[Kyunghyun: #1]}}

\maketitle

\begin{abstract}
Statistical inference, a central tool of science, revolves around the study and the usage of statistical estimators: functions that map finite samples to predictions about unknown distribution parameters. 
In the frequentist framework, estimators are evaluated based on properties such as bias, variance (for parameter estimation), accuracy, power, and calibration (for hypothesis testing). However, crafting estimators with desirable properties is often analytically challenging, and sometimes impossible, e.g., there exists no universally unbiased estimator for the standard deviation.
In this work, we introduce meta-statistical learning, an amortized learning framework that recasts estimator design as an optimization problem via supervised learning. This takes a fully empirical approach to discovering statistical estimators; entire datasets are input to permutation-invariant neural networks, such as Set Transformers, trained to predict the target statistical property. The trained model is the estimator, and can be analyzed through the classical frequentist lens.
We demonstrate the approach on two tasks: learning a normality test (classification) and estimating mutual information (regression), achieving strong results even with small models. Looking ahead, this paradigm opens a path to automate the discovery of generalizable and flexible statistical estimators.
\end{abstract}

% \Megan{
% \begin{enumerate}
%     \item @Megan if it's not difficult, please increase the x, y-ticks and labels on the consistency graphs. \Megan{Yes, apologies!}
% \end{enumerate}
% }

\section{Introduction}
\label{sec:introduction}
\input{01_intro}

\section{Meta-Statistical Learning}
\label{sec:meta_ml}
\input{03_framework}

\section{Experiments}
\label{sec:experiments_inf}
\input{04_experiments}

\section{Discussion}
\label{sec:discussion}
\input{05_discussion}
\section*{Broader Impact Statement}
This paper presents work whose goal is to advance the field of Machine Learning. Meta-statistical learning aims to enhance inference in low-sample settings, benefiting applied fields of Science like medicine and economics by improving estimator reliability. Learned estimators may inherit bias from the data they would be trained on, potentially leading to misleading conclusions if not carefully validated. Further, as with any data-driven methodology, interpretability remains a challenge; understanding why a model makes a particular statistical inference is crucial for scientific rigor. 

\section*{Acknowledgments}
\input{11_acknowledgements}

\clearpage

\bibliography{iclr2026_conference}
\bibliographystyle{iclr2026_conference}

\appendix
\input{10_appendix}

\end{document}

%% file: math_commands.tex
\usepackage{amsmath,amsfonts,bm}

\def\Figref#1{Figure~\ref{#1}}

\def\Secref#1{Section~\ref{#1}}
\def\eqref#1{equation~\ref{#1}}
\def\1{\bm{1}}

\DeclareMathAlphabet{\mathsfit}{\encodingdefault}{\sfdefault}{m}{sl}
\SetMathAlphabet{\mathsfit}{bold}{\encodingdefault}{\sfdefault}{bx}{n}

\newcommand{\E}{\mathbb{E}}

%% file: 01_intro.tex
Statistical inference is the principled approach to quantifying uncertainty, testing hypotheses, and drawing conclusions from data~\citep{walker1953statistical,casella2024statistical}. It is central to scientific inquiry across disciplines, guiding the design of experiments, the validation of theories, and the interpretation of empirical results~\citep{barlow1993statistics, altman1990practical, james2006statistical, dienes2008understanding, salganik2019bit}. Yet, in practice, statistical inference is notoriously challenging. Real-world data often deviates from idealized assumptions~\citep{gurland1971simple, hoekstra2012assumptions,knief2021violating,NEURIPS2023_36b80eae}, and inference in low-sample regimes presents a persistent challenge.
Recent advances in machine learning offer tools to assist the practice of statistical inference. For example, the costly computation of the Bayes rule can be amortized over datasets sampled within a fixed data simulator by training a neural network to approximate the posterior distribution~\citep{pmlr-v89-papamakarios19a, cranmer2020frontier, simons2023neural}. 
This perspective originating in \textit{simulation-based inference}~\citep{cranmer2020frontier} gave rise to the active field of research known as \textit{amortized Bayesian inference}~\citep{chan2018likelihood, chen2023learning, chen2020neural, radev2020bayesflow, avecilla2022neural, pmlr-v235-gloeckler24a}.

In this work, we take a different path and focus on the frequentist concept of \textbf{statistical estimators}, which are functions mapping finite samples to predictions of parameters of unknown, but fixed, distributions~\citep{fisher1925theory}. Classically, estimators are evaluated via properties like \textit{bias}, \textit{variance}, \textit{calibration}, or \textit{power}~\citep{cox2006principles}, and are often derived analytically to satisfy guarantees uniformly over a class of distributions $\Gamma$. For example, we might look for the \textit{minimum variance unbiased estimator} (MVUE) of some quantity $\theta$ over $\Gamma$. However, analytical derivation of desirable estimators can be challenging or even infeasible. For example, no universally unbiased estimator of the standard deviation exists~\citep{gurland1971simple,bengio2003no}.

We introduce \emph{meta-statistical learning}, a framework that recasts the design of statistical estimators as an optimization problem solved via supervised learning. Neural estimators are implemented via permutation-invariant architectures such as Set Transformers~\citep{lee2019set}, enabling predictions over variable-sized exchangeable sequences. 
They can then be optimized for and evaluated against any desired frequentist property or combination thereof, such as bias-variance trade-offs or power.
During training, we must sample a collection of datasets drawn from the set of possible data-generating distributions $\Gamma$. However, since it is infeasible to sample \emph{uniformly} from $\Gamma$, we are forced to introduce a meta-prior $P_\Gamma$. Therefore, we focus on \textbf{out-of-meta-distribution} generalization: scenarios where the meta-prior shifts at test time. This shift selects meta-statistical estimators that are \textit{more invariant} to the choice of meta-prior.

\paragraph{Contributions.}
We (i) introduce the learning of statistical estimators trained to optimize relaxed differentiable frequentist properties; (ii) evaluate various architectures for efficient dataset encoding and generalization; and (iii) demonstrate the approach on two general tasks: learning a normality test and learning mutual information estimators, achieving strong performance even with tiny models. (iv) We release our code and model.\footnote{\url{https://github.com/PeyrardM/meta-statistical-learning}. For mutual information estimation, also check \url{https://github.com/grgera/mist}, from dedicated follow-up work~\citep{gritsai2025mistmutualinformationsupervised}.}

Looking ahead, this approach contributes to the larger project of meta-learning statistical properties and opens avenues for automating the discovery of generalizable statistical estimators.

%% file: 03_framework.tex
Suppose a family of data‐generating distributions $\Gamma$ is given, where each $\gamma\in\Gamma$ is a probability law over an observation space $\mathcal{X}$. A statistical functional $\theta = g(\gamma)\in\Theta$ extracts the parameter of interest from $\gamma$.  To estimate $\theta$, we observe $n$ i.i.d.\ samples $X_1,\dots,X_n\sim\gamma$, forming a dataset $X\in\mathcal{X}^n$, and apply an estimator $f:\bigcup_{n=1}^\infty\mathcal{X}^n\to\Theta$ to yield $\hat\theta=f(X)$.  In the classical setting, one designs $f$ analytically to satisfy frequentist criteria such as unbiasedness, consistency, or minimum variance over all $\gamma\in\Gamma$, e.g.\ the sample mean is the minimum variance unbiased estimator (MVUE) for the population mean of distributions within the Gaussian family but not for the Uniform family. Yet, in many scenarios, the analytical discovery of an estimator can be challenging or even impossible.

\begin{figure}
    \centering
    \includegraphics[width=0.9\linewidth]{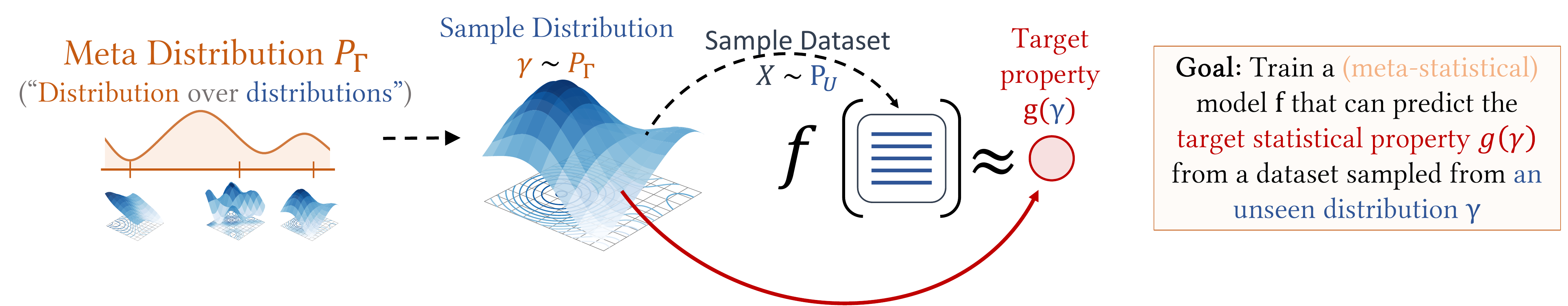}
    \caption{Illustration of Meta-Statistical Learning}
    \label{fig:fig1}
\end{figure}

We instead propose to learn $f$ by optimizing over a parameterized class $\mathcal{F}$ to minimize a candidate risk $\mathcal{L}\bigl(f(X),\theta\bigr)$ in expectation. Then, we can interpret an approximation of frequentist estimators design as solving:
% \begin{equation}
% \arg\min_{f\in\mathcal{F}}\;\sup_{\gamma\in\Gamma}\;\mathbb{E}_{X\sim\gamma}\bigl[\mathcal{L}\bigl(f(X),g(\gamma)\bigr)\bigr].
% \end{equation}
% However, in practice we have to sample from $\Gamma$ which requires the specification of some meta-prior $P_{\Gamma}$ over $\Gamma$ leading to a relaxed objective
\begin{equation}
\arg\min_{f\in\mathcal{F}}\;\mathbb{E}_{\gamma\sim P_{\Gamma}}\;\mathbb{E}_{X\sim\gamma}\bigl[\mathcal{L}\bigl(f(X),g(\gamma)\bigr)\bigr].
\end{equation}
Here $P_{\Gamma}$ enables training by sampling distributions, but it can make $f$ favor regions of higher meta‐prior density. 
To better approach the frequentist objective, we aim to generalize beyond the meta-prior. To do so, we apply various strategies: (i) validating and selecting estimators under \textbf{out‐of‐meta‐distribution} priors, (ii) incorporating losses that encourage desirable frequentist behavior, and (iii) expanding $P_{\Gamma}$ to be as uninformative and diverse as possible (scaling training data). In future work, training strategies like distributionally robust~\citep{rahimian2019distributionally} or invariant techniques~\citep{arjovsky2020invariantriskminimization} could also be explored. We now split the discussion of this estimator‐learning paradigm into two settings: real‐valued $\theta$'s (regression) and hypothesis tests (classification).

% To mitigate this issue, we introduce several strategies:
% \begin{itemize}
%     \item \textbf{Loss augmentation.} We modify the loss function to explicitly encourage desirable frequentist properties such as unbiasedness and low variance, even if this comes at a cost to the standard expected error under $P_{\Gamma}$.
%     \item \textbf{Out-of-meta-distribution model selection.} We validate model performance on held-out distributions that lie outside the support of $P_{\Gamma}$, thereby favoring estimators that generalize beyond the training prior.
%     \item \textbf{Diverse sampling.} We aim to sample distributions from $P_{\Gamma}$ as diversely as possible to encourage robustness to variation in $\gamma$ and reduce overfitting to narrow distributional regimes.
% \end{itemize}

% In future work, we plan to explore alternative approaches for mitigating the effect of non-uniform priors. This includes experimenting with uninformative priors, or adapting techniques from domain adaptation and distribution shift literature, such as methods designed to address label shift (changes in $P(y)$ under fixed $P(X|y)$) when estimating $P(y|X)$).

\subsection{Parameter Estimation: Meta-Statistical Regression}
We now specialize to the case where the parameter of interest is real-valued  \( (\Theta = \mathbb{R}) \). Typically, we want to analyze the behavior of an estimator \( f \) under the frequentist notions of bias and variance, which are defined for a fixed $\gamma$:
\begin{align}
    \text{Bias}_\gamma(f) &= \mathbb{E}_{X \sim \gamma}[f(X)] - g(\gamma), \\
    \text{Var}_\gamma(f) &= \mathbb{E}_{X \sim \gamma} \left[ \left( f(X) - \mathbb{E}_{X \sim \gamma}[f(X)] \right)^2 \right].
\end{align}

To learn an estimator with both low bias and variance, an first idea is to minimize $\text{MSE}_{\gamma}(f) = \text{Bias}^2_\gamma(f) + \text{Var}_\gamma(f) $, and average over $\gamma \sim P_\Gamma$. In fact, this simply corresponds to training the meta-statistical model to perform MSE regression on the full meta-dataset (see Appendix~\ref{app:math_regr}):
\begin{align}
    \mathcal{L}_{\text{MSE}}(f) &= \mathbb{E}_{\gamma \sim P_\Gamma} \mathbb{E}_{X \sim \gamma} \left[ (f(X) - g(\gamma))^2 \right] \\
    &= \mathbb{E}_{\gamma \sim P_\Gamma} \left[ \text{Bias}_\gamma(f)^2 + \text{Var}_\gamma(f) \right].
\end{align}
The first expression defines a regression objective on the meta-dataset. The second expression reveals that minimizing the MSE corresponds to minimizing the average (over $P_\Gamma$) frequentist bias and variance of \( f \) as a statistical estimator.

\noindent \textbf{Towards MVUE.} 
In the limit of perfect optimization and sufficient model capacity, the minimizer of \( \mathcal{L}_{\text{MSE}} \) is the Bayes regression function:
\begin{equation}
f^*(x) = \mathbb{E} \left[ g(\gamma) \mid X = x \right],
\end{equation}
which represents the posterior expectation of \( g(\gamma) \) given the observed dataset \( X = x \) under the prior distribution \( P_\Gamma \). This estimator is sensitive to the choice of meta-prior \( P_\Gamma \). However, in our setting, \( P_\Gamma \) is not intended to encode meaningful beliefs. Rather, it is an artifact of the practical constraint that we cannot uniformly sample from \( \Gamma \). 
% In fact, \( f^*(x) \) is typically biased for any fixed \( \gamma \in \Gamma \).

Alternatively, one can seek to approximate the minimum variance unbiased estimator (MVUE) when it exists. To this end, we can formulate a constrained optimization problem:
\begin{equation}
    \min_{f} \; \mathbb{E}_{\gamma \sim P_\Gamma} \left[ \text{Var}_{X \sim \gamma}(f(X)) \right] \quad \text{subject to} \quad \mathbb{E}_{\gamma \sim P_\Gamma} \left[ \text{Bias}_\gamma(f)^2 \right] = 0.
\end{equation}
The constraint enforces that the estimator is unbiased on average across \( \gamma \sim P_\Gamma \). Using a Lagrangian relaxation, we arrive at the following loss expression:
\begin{equation}
\mathcal{L}_{\text{MVUE}}(f) = \mathcal{L}_{\text{MSE}}(f) + \lambda \cdot \mathbb{E}_{\gamma \sim P_\Gamma} \left[ \text{Bias}_\gamma(f)^2 \right],
\end{equation}
where \( \lambda > 0 \) is a factor controlling the strength of the bias penalty. Since \( \mathcal{L}_{\text{MSE}} \) already decomposes into the conditional frequentist variance and squared bias, this loss corresponds to increasing the penalty on bias. In practice, the bias penalty term can be estimated by a batch-level empirical bias.
% \[
% \left( \frac{1}{k} \sum_{i=1}^k f(X'_i) \right)^2,
% \]
% which is weighted by the hyperparameter \( \lambda \) and added to the batch-level MSE loss.

\noindent \textbf{Relationship to Amortized Bayesian Inference (ABI)} 
The meta-statistical approach is a learning problem: predicting $\theta = g(\gamma)$ from data samples $X$ under a sampling strategy dictated by the meta-prior $P_{\Gamma}$. A very intuitive idea would be to replace point-estimates made for $\theta$ by posterior distributions $p(\theta \mid X)$. Since we have access to simulator $P_{\Gamma}$ that generates data $(\gamma, X)$, and computing Bayes rule to get the posterior is intractable, it appears we are in a perfect setting for \textit{amortized Bayesian inference} \citep{chan2018likelihood, chen2023learning, chen2020neural, elsemueller2024sensitivity, radev2020bayesflow, avecilla2022neural, pmlr-v235-gloeckler24a} to learn a parametrized model $q(\theta \mid X)$. Training is done with generative learning to approximate $p(\theta|X)$:
\begin{equation}
    \mathcal{L}_{\text{ABI}}(q) = \mathbb{E}_{\gamma \sim p_\Gamma} \, \mathbb{E}_{X \sim \gamma} \left[ -\log q(\theta \mid X) \right].
\end{equation}
The connection between ABI and meta-statistical learning is rich. In our setting, the estimator learns a point estimates and optimizes frequentist properties. ABI has a similar setup except it learns to output posterior distributions and is framed as a generative loss. Taking expectations of ABI's posteriors recovers a point estimates. Conversely, the meta-statistical learning can approximate ABI by, for instance, learning to predict the quantile of the posterior distribution (e.g.,  with a pinball loss~\citep{pinball_loss,NEURIPS2019_73c03186} or the $L_{NNL}$ described above). The important conceptual difference lies in the treatment of the posterior and the frequentists vs Bayesian perspective: ABI assumes a fixed prior, purposefully models it; the posterior approximation depends on it and will not generalize to new priors. In contrast, our framework aims to generalize across data-generating distributions, removing its dependence on the prior. Also, the target quantity does not need to have any special connection with a posterior, it could be any functional of the distribution. This conceptual difference means that meta-statistical learning strives for uninformative priors, a regime where ABI will have large variance, which we observe empirically in our experiments (\Secref{sec:mi_exp}).

% However, our problem is better understood as not one simulator following $P_{\Gamma}$ but many simulators (one for each $\gamma \in \Gamma$). The meta-prior $P_{\Gamma}$ over $\gamma$, which induces the prior over $\theta$, does not reflect useful prior belief but is the undesired by-product of sampling; we want to generalize across different meta-priors. ABI purposefully models the meta-prior; its posterior approximation depends on it and will not generalize to new meta-priors. This is what we observe in \Secref{sec:mi_exp}.

\noindent \textbf{Modeling Predictive Uncertainty.}
One powerful insight from ABI is the importance of modeling predictive uncertainty, which is not built-in into point estimators. Rather than approximating the full posterior over the meta-prior $P_\Gamma$, the frequentist perspective would require modeling prediction uncertainties for resampled $X$ from a fixed $\gamma$. In particular, one desired frequentist criterion is asymptotic normality: the difference between predictions and the true parameter $\theta$ should converge \emph{in distribution} to a normal distribution and the variance of errors should approach the Cramér-Rao bound
$\text{Var}_{\gamma}(f) \geq \frac{1}{n}\mathcal{I}^{-1}_{\gamma}(\theta)$, 
where $\mathcal{I}_{\gamma}(\theta)$ is the Fisher information of the distribution $\gamma$ for the parameter $\theta$. 
The fundamental unpredictability of $\theta$ from $X$ according to the best possible estimator (achieving the Cramér-Rao bound) depends on $\frac{1}{n}\mathcal{I}^{-1}_{\gamma}(\theta)$, which varies for different distributions and varies (decreases) for resamples of different sizes of the same distribution. This conflicts with the implicit homoscedasticity assumption of MSE: constant irreducible noise variance across inputs. 

We can easily accommodate the Gaussian shape and heteroscedastic regression by optimizing a Gaussian negative log-likelihood loss:
\begin{equation}
\mathcal{L}_{\text{NLL}}(f, s) = \mathbb{E}_{\gamma \sim \Gamma} \, \mathbb{E}_{X \sim P_\gamma} \left[ \frac{1}{2} \log s(X) + \frac{(f(X) - g(\gamma))^2}{2s(X)} \right],
\end{equation}
where the model outputs both the mean $f(X)$ but also an input-dependent variance estimate $s(X)$. Interestingly, this also corresponds to the default variational approximation to amortized Bayesian inference \citep{wu2020meta,ganguly2023amortized,margossian2024amortizedvariationalinferencewhy}, which is here minimized in expectation across $\gamma \in \Gamma$.

\subsection{Hypothesis Testing: Meta-Statistical Classification}
Let $\Theta = \{0, 1\}$ denote a binary variable corresponding to a statistical property of the underlying data-generating distribution $\gamma$. We partition the space of distributions $\Gamma$ into two disjoint subsets: $\Gamma_0$ (null hypothesis) and $\Gamma_1$ (alternative hypothesis). These subsets are associated with meta-priors $P_0$ and $P_1$. Given a sample $X \sim \gamma$ for some $\gamma \in \Gamma$, the task is to infer $\theta = \mathbb{I}[\gamma \in \Gamma_1]$.

In this framework, a learned function $f$ that maps samples $X$ to the interval $[0,1]$ is a (meta-statistical) hypothesis test. The canonical training objective to maximize the accuracy for such a classifier is the binary cross-entropy loss, defined as:

\begin{align}
\mathcal{L}_{\mathrm{CE}}(f) &= 
\mathbb{E}_{\gamma \sim P_1} \mathbb{E}_{X \sim \gamma} \left[-\log f(X) \right]
+
\mathbb{E}_{\gamma \sim P_0} \mathbb{E}_{X \sim \gamma} \left[-\log \left(1 - f(X) \right) \right], \\
&= - \mathbb{E}_{\gamma \sim P_\Gamma} \mathbb{E}_{X \sim \gamma} \left[ \theta \cdot \log f(X) + (1 - \theta) \cdot \log(1 - f(X))\right].
\end{align}

This loss minimizes the expected Kullback--Leibler divergence between the true posterior $p(\theta \mid X)$ and the model's prediction $f(X)$, recovering the Bayes-optimal classifier in the limit. A classifier $f$ trained with $\mathcal{L}_{CE}$ approximates the global posterior under the meta-distribution, as if performing amortized Bayesian inference (ABI) with the global simulator.
Unlike the regression case, where defining a uniform prior over $\Gamma$ is infeasible, we can easily impose a uniform prior over $\theta$ by sampling $\gamma$ equally from $\Gamma_0$ and $\Gamma_1$. However, achieving uniformity within each family $\Gamma_0$ and $\Gamma_1$ remains infeasible in general.

\noindent \textbf{Neyman--Pearson Loss.}
While the Bayes classifier is optimal with respect to overall classification accuracy, it does not explicitly optimize other frequentist properties such as power under a fixed Type I error rate \citep{sciadv.aao1659}.
Given a learned score function $f : X \mapsto \mathbb{R}$, a test is induced by comparing $f(X)$ to a threshold $t$. In the previous cross-entropy loss, the threshold is fixed to $t=0.5$. For a fixed distribution $\gamma$, the resulting Type~I and Type~II errors are
\[
\alpha_\gamma(f,t) = \mathbb{P}_{X\sim\gamma}\!\left[f(X) > t\right], 
\qquad \gamma \in \Gamma_0,
\]
\[
\beta_\gamma(f,t) = \mathbb{P}_{X\sim\gamma}\!\left[f(X) \le t\right],
\qquad \gamma \in \Gamma_1,
\]
with corresponding power $1-\beta_\gamma(f,t)$.
The Neyman--Pearson objective seeks the most powerful test under a Type~I error constraint $\alpha_0$:
\begin{equation}
\min_{f,t}\; 
\mathbb{E}_{\gamma\sim P_1}\!\left[\beta_\gamma(f,t)\right]
\quad\text{s.t.}\quad
\mathbb{E}_{\gamma\sim P_0}\!\left[\alpha_\gamma(f,t)\right] \le \alpha_0.
\label{eq:np_constraint}
\end{equation}

We can estimate $\beta_{\gamma}$ and $\alpha_\gamma$ with a batch of training data, by counting the true positive rate and false positive rate in the batch. Since counting is not differentiable, we replace $\mathbf{1}\!\{\,f(X) > t\,\}$ with a smooth approximation with the sigmoid function $\sigma$:
% Relaxing the constraint in~\eqref{eq:np_constraint} with a Lagrange multiplier $\lambda>0$ yields the unconstrained objective
% \begin{equation}
% \min_{f,t}
% \;\mathbb{E}_{\gamma\sim P_1}\!\left[ 1 - f_t(X) \right]
% \;+\;
% \lambda\, \mathbb{E}_{\gamma\sim P_0}\!\left[ f_t(X) - \alpha_0 \right],
% \label{eq:np_lagrangian}
% \end{equation}
% where $f_t(X)=\mathbf{1}\!\{\,f(X) > t\,\}$ is the rejection indicator. Since $f_t(X)$ is non-differentiable, we replace it with the smooth approximation
\[
\sigma_t(X)
=
\sigma\!\left(\frac{f(X)-t}{\tau}\right),
\]
where $\tau>0$ controls the sharpness of the surrogate.\footnote{To stabilize training and avoid degenerate shifts in $f$, we standardize the scores within each batch,
$\hat{f}(X) = \frac{f(X) - \mu}{\sigma},$
expressing the threshold $t$ in standardized units. }
Both $f$ and $t$ are then optimized jointly, and $\tau$ is an hyper-parameter.

% Then, we obtain the following loss:
% \begin{equation}
% \mathcal{L}_{\mathrm{NP}}
% = \mathbb{E}_{\gamma \sim P_{\Gamma}} \mathbb{E}_{X \sim \gamma} \left[ \theta \cdot \left(1 - h_t(X) \right) + \lambda \cdot \left(1 - \theta \right) \cdot \left(h_t(X) - \alpha_0\right) \right]
% \end{equation}

Then, for a batch of data $\{X_i, y_i\}_{i \in 1 \dots B}$ of size $B$, the training loss becomes:
\begin{equation}
\mathcal{L}_{\mathrm{NP}}^{\text{batch}}
= \left(1 - \underbrace{\E_{B} \left[\sigma_t(X)|y=1\right]}_{\text{TPR (power) in batch}} \right)
+
\lambda \, \,\max\!\left(0,\; \underbrace{\E_{B} \left[\sigma_t(X)|y=0\right]}_{\text{FPR in batch}} - \alpha_0\right)
\label{eq:pc_loss}
\end{equation}
The first part of the loss consists in maximizing the true positive rate (TPR) which is power, and the second part consists in keeping the false positive rate (FPR) under threshold $\alpha_0$, if the FPR goes below $\alpha_0$, no loss is incurred. 
The function $\max(0, \cdot)$ is implemented with a ReLU. This provides a differentiable relaxation of the Neyman--Pearson criterion~\citep{plugin_np, sciadv.aao1659}, enabling end-to-end training while enforcing a desired Type~I error level.

\subsection{Related Work}
Processing multiple data points simultaneously originates in \textit{multi-instance learning}, where models receive sets of instances and assign labels at the group level \citep{maron1997framework, dietterich1997solving, ilse2018attention,NEURIPS2022_ac56fb3f}. A relevant example is the \textit{neural statistician} framework \citep{edwards2017towards}, which employs variational autoencoders (VAEs) to learn dataset representations in an unsupervised manner \citep{hewitt2018variational}. Learning dataset-level representations has also been explored through meta-features \citep{jomaa2021dataset2vec, kotlar2021novel, 10136150, wu2022learning}, where models extract high-level statistics tailored for specific tasks. Recently, \citet{hollmann2025accurate} employed transformers trained on synthetic datasets for missing value imputation, which we recognize as an instance of meta-statistical learning in low-sample-size settings.
Approaches of a meta-statistical nature have also been successfully applied in causal discovery \citep{lopez2015towards, lowe2022amortized, lorch2022amortized, wu2024sample}. These methods generate synthetic data with known causal structures and train neural networks to infer causal properties from a set of observations \citep{kelearning,kim2024targeted}.

\noindent \textbf{Machine Learning for Statistical Inference.}
Our work aligns with the broader research direction on neural processes \citep{garnelo2018neural, garnelo2018conditional,kim2019attentive, Gordon:2020, markou2022practical, huang2023practical, BruinsmaMRFAVBH23}. Neural processes can predict latent variables of interest from datasets \citep{chang2025amortized} by leveraging transformers \citep{pmlr-v162-nguyen22b} and Deep Sets \citep{NIPS2017_f22e4747} to enforce permutation invariance \citep{JMLR:v21:19-322}. A related approach, known as prior-fitted networks, has demonstrated that transformers can be effectively repurposed for Bayesian inference \citep{muller2022transformers} and optimization tasks \citep{pmlr-v202-muller23a}.
Additionally, there is growing interest in using trained models to assist in statistical inference tasks \citep{angelopoulos2023predictionpoweredinference} and optimization \citep{NIPS2017_addfa9b7, NEURIPS2020_f52db9f7, NEURIPS2021_56c3b2c6, amos2023tutorialamortizedoptimization} like  
% \citep{angelopoulos2023predictionpoweredinference} and optimization \citep{NIPS2017_addfa9b7, NEURIPS2020_f52db9f7, NEURIPS2021_56c3b2c6, amos2023tutorialamortizedoptimization}. In particular, 
simulation-based inference (SBI) \citep{pmlr-v89-papamakarios19a, cranmer2020frontier} and amortized Bayesian inference (ABI) \citep{gonccalves2020training, elsemueller2024sensitivity, radev2020bayesflow, avecilla2022neural, pmlr-v235-gloeckler24a} 
% typically replaces probabilistic inference with a neural network prediction task \citep{chan2018likelihood, chen2023learning, chen2020neural}. 
As discussed previously, SBI and ABI approaches if applied naively to our setup model the unwanted meta-prior. 
Recent works use regression tasks to probe in-context learning in decoder-only transformers \citep{bai2023transformers, song2025decodingbasedregression, nguyen2024predictingstringslanguagemodel}. For example, \cite{reuter2025transformerslearnbayesianinference} learns Bayesian inference from decoder Transformer architectures. 
% Indeed, attention-based models are well suited to represent exchangeable input datasets. SetTransformer-based encoders \citep{lee2019set, zhang2022set} address scalability by reducing attention costs and naturally enforce permutation invariance. 
Our approach builds uses encoder-only SetTransformers \citep{lee2019set, zhang2022set} trained via supervised learning. Finally, close to the tasks of this paper, \citet{simić2020testingnormalityneuralnetworks} trains a multi-layer perceptron for normality testing only for fixed sized datasets and fixed dimensions, and \citet{hu2024infonetneuralestimationmutual} proposes InfoNet a meta-learned mutual information estimator.

\noindent \textbf{Meta-Learning.}
Meta-learning aims to generalize across tasks by acquiring transferable knowledge \citep{schmidhuber1996simple, schmidhuber1987evolutionary, thrun1998lifelong, schmidhuber1993neural, vanschoren2019meta, hospedales2021meta, huisman2021survey}. Dataset-level encoders have been explored in few-shot learning \citep{vinyals2016matching, santoro2016meta, finn2017model, snell2017prototypical, wang2023few, wu2020meta, rivolli2022meta, mishra2017simple, ravi2017optimization, munkhdalai2017meta, shyam2017attentive}. Neural processes, for example, learn function priors that adapt to new datasets using observed data \citep{garnelo2018neural, garnelo2018conditional, kim2019attentive}. Meta-statistical learning shares similar goals of generalization across priors and could thus import methods from meta-learning.

%% file: 04_experiments.tex
To validate our framework, we design a two-stage experimental protocol that moves from controlled settings to meaningful statistical tasks.
First, we focus on architecture selection, using dataset-level tasks that do not involve statistical uncertainty. These tasks serve two purposes: they allow us to compare permutation-invariant encoders in a clean environment, and they demonstrate that our framework naturally accommodates amortised optimization and algorithm learning. This stage produces the architecture we rely on for subsequent experiments.
Second, using the selected architecture, we evaluate the framework on two meaningful statistical inference problems: (i) learning a normality test (classification) and (ii) estimating mutual information (regression). While this paper does not aim to solve each of these problem, these two tasks cover both classification and regression while also representing challenging inference settings. The strong results provide concrete evidence that the framework can learn non-trivial statistical inference and highlight the potential for future large-scale efforts toward foundation models of statistical inference.

To construct the test meta-datasets, we employ a grid-based sampling strategy. For each distribution family in the test meta-distribution, we sample $m$ distinct parameterizations. Then, for each dataset size in a predefined grid spanning $[10, 300]$, we generate $k$ datasets from each parameterization. This repeated resampling enables us to assess the frequentist behaviour of the learned estimators as a function of dataset size and distribution families.
Finally, to assess out-of-meta-distribution (OOMD) generalization, we modify the support of $P_\Gamma$ by adding distribution families never observed during training.

\subsection{Selection of a Meta-Statistical Architecture}
\label{sec:arch_choice}
Meta-statistical models must handle variable-sized input datasets \( X \in \mathbb{R}^{n \times k} \). Therefore, we use a permutation-invariant encoder \( \phi \) followed by a prediction head \( \rho \), typically a multi-layer perceptron. The model \( f(X) = \rho \circ \phi(X) \) is trained to minimize task-specific losses, encouraging \( \phi \) to learn sufficient statistics of \( X \) for the inference task.
We compare several encoder architectures on a set of descriptive tasks (e.g., per-column median, optimal transport, win-rate, \dots). As a baseline, we include an LSTM \citep{hochreiter1997long}, which lacks permutation invariance. The Vanilla Transformer (VT) and the Set Transformer \citep{lee2019set}, which reduces the quadratic cost of attention by introducing a fixed number of learned inducing points on which attention is performed. We adopt an improved version, Set Transformer 2 (that we call ST here) \citep{zhang2022set}, which uses SetNorm \citep{ba2016layer} to preserve permutation invariance while improving empirical training convergence.

Here, we briefly summarize the results, but details on generation, training, and evaluation metrics are documented in Appendix~\ref{app:desc_details}.
Results show that both VT and ST generalize well to unseen distributions and larger datasets than seen during training. LSTM underperforms, which emphasizes the value of permutation invariance. ST with 32 inducing points is roughly six times faster per batch than VT and scales much better at inference time (as expected). For all subsequent experiments, we use ST with 32 inducing points as the default encoder.

\subsection{Supervised Learning of Normality Tests}
\label{sec:norm_exp}
Normality testing is crucial in practical inference pipeline\citep{shapiro1965analysis, razali2011power}, particularly before applying t-tests, linear regression, or ANOVA with small samples \citep{altman1990practical, das2016brief, doi:10.4078/jrd.2019.26.1.5}. This offers a nice and challenging test-bed for meta-statistical learning. 
We construct a balanced meta-dataset of normally and not normally distributed datasets. For non-normality, we draw non-normal distributions from a predefined set, keeping unseen distribution families for OOMD testing. As it is common practice, we standardize every dataset $X$. The training meta-dataset has $2M$ pairs.
Additional details about the meta-dataset, model properties, and training are in Appendix~\ref{app:norm_test}.

\begin{figure}
    \centering
    \includegraphics[width=\linewidth]{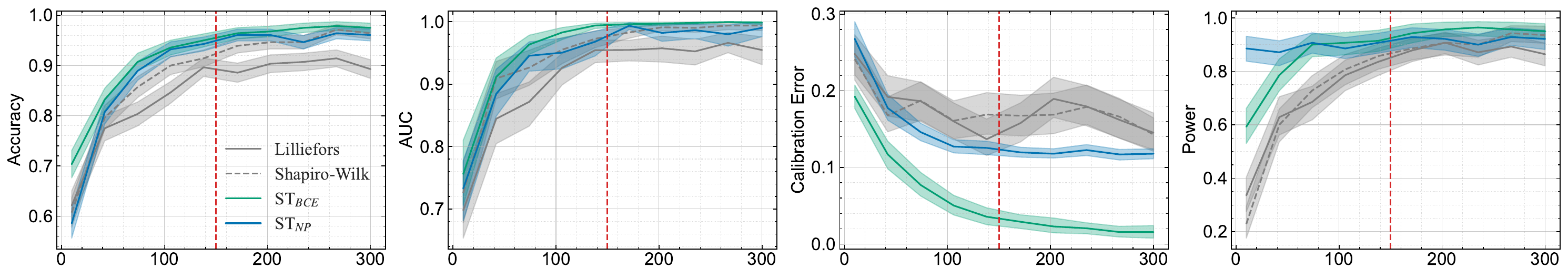}
    \caption{\textbf{OOMD evaluation} of normality tests as a function of the input dataset size against the two best baselines: Shapiro-Wilk and Lilliefors. (Red line: training cut-off)}
    \label{fig:norm_results}
\end{figure}

\paragraph{Estimators Compared.}  
As baselines, we consider six widely used standard normality tests: the \textit{Shapiro-Wilk test} \citep{shapiro1965analysis}, known to be effective for small samples \citep{razali2011power}; the \textit{D'Agostino-Pearson test} \citep{d1973tests}, which combines skewness and kurtosis; the \textit{Kolmogorov-Smirnov test} \citep{massey1951kolmogorov}, a non-parametric test based on cumulative distribution differences; the \textit{Jarque-Bera test} \citep{jarque1987test}, which assesses skewness and kurtosis deviations from theoretical expectations; the \textit{Lilliefors test} \citep{lilliefors1967kolmogorov}, a generalization of the Kolmogorov-Smirnov test; and the \textit{Anderson-Darling test} \citep{anderson1952asymptotic}, which is based on the empirical distribution function and does not output p-values relies on \textit{critical values}. In the main paper, for clarity, we report only the two best baselines (Lilliefors and Shapiro-Wilk), the full results are available in Appendix~\ref{app:norm_test}.
We then train and evaluate two meta-statistical tests sharing the same ST architecture (140K parameters): \textbf{ST\textsubscript{BCE}} trained with the binary cross-entropy loss, \textbf{ST\textsubscript{NP}} trained with the Neyman-Pearson loss ($\tau=0.1$ and $\alpha_0=0.05$).

\paragraph{Evaluation Protocol.}
We evaluate each normality test by measuring its \textbf{accuracy}, its Area Under the Receiver Operating Characteristic Curve (\textbf{AuROC}), its \textbf{calibration}, and its \textbf{power} at level $0.05$. The accuracy requires a choice of threshold on the estimate $p(\gamma \in \Gamma_1)$, which is naturally set to $0.5$ for the learned models. To measure accuracy of baseline normality tests, we optimize their p-value thresholds (or critical values) independently on the same validation sets as the ones used to train the meta-statistical tests. The AUROC measures a classifier's ability to discriminate between positive and negative classes across different decision thresholds. The calibration error measures the mismatch between predicted probabilities and observed error rates. Finally, the power measures the ability to recognize the alternative hypothesis (non-normal) when it is there, we fix the threshold $\alpha=0.05$ for comparable results. The power and the calibration error are computed based on $k=100$ resamples from each distribution in the test meta-datasets and averaged over all distributions. To evaluate generalization beyond the training regime, we assess the consistency of our estimators by tracking their metrics as a function of input size \(n\) in \Figref{fig:norm_results}.

\paragraph{Results and Analysis.}
In this OOMD setting, meta-statistical tests outperform standard ones, including extrapolation to dataset sizes unseen during training. Interestingly, the NP loss behaves as expected; it significantly increases the power of the learned test compared to the BCE loss. The results show that frequentist desiderata can be achieved by simple loss design. The learned estimators are particularly well-calibrated (especially BCE), meaning their predicted probabilities $f(X) \approx P(\gamma \in \Gamma_1 | X)$ accurately reflect the uncertainty from $X$.

\subsection{Supervised Learning of Mutual Information Estimators}
\label{sec:mi_exp}
Mutual information (MI) quantifies the dependence between two random variables \(X\) and \(Y\).
In the continuous form, it is defined as:
\begin{equation}
\mathrm{MI}(X; Y) = \int \int P_{X,Y}(x, y) \log \frac{P_{X,Y}(x, y)}{P_X(x) P_Y(y)} \, dx \, dy.
\end{equation}
Here, \(P_X\) and \(P_Y\) denote the marginal distributions of \(X\) and \(Y\), respectively. 
MI displays key properties such as invariance under homeomorphisms and adherence to the \textit{Data Processing Inequality}, making it a foundational quantity in machine learning \citep{inv_bottleneck, pmlr-v80-belghazi18a, repr_learning, tishby2000informationbottleneckmethod}. However, its estimation remains difficult, particularly in regimes with limited sample sizes or non-Gaussian distributions \citep{song2020understandinglimitationsvariationalmutual, pmlr-v108-mcallester20a, NEURIPS2023_36b80eae}. We adopt a meta-statistical perspective, training models to predict \( \theta = \mathrm{MI}(X; Y) \).

\begin{figure}[ht]
    \centering
    \begin{minipage}[t]{0.42\textwidth}
        \vspace{0pt} % ensure top alignment
        \includegraphics[width=\linewidth]{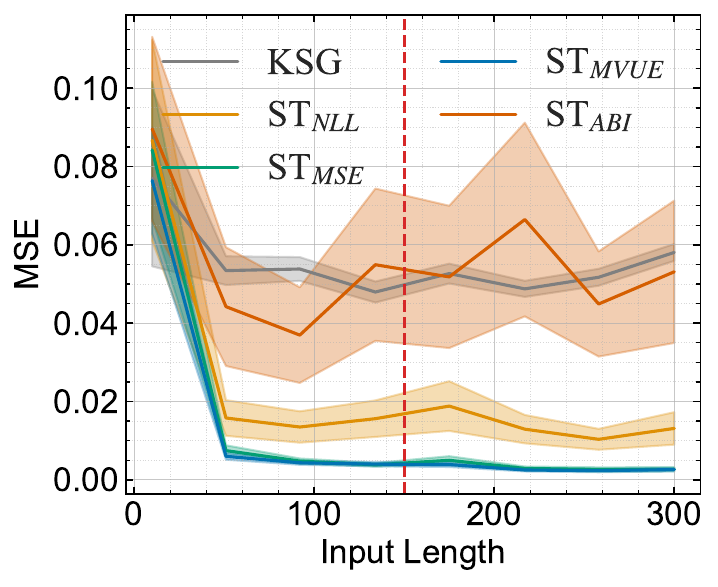}
        \caption{\textbf{OOMD consistency} of MI estimators: MSE as a function of the input dataset size. (Red line: training cut-off)}
        \label{fig:mi_consistency}
    \end{minipage}%
    \hfill
    \begin{minipage}[t]{0.54\textwidth}
        \vspace{0pt} % ensure top alignment
        \captionof{table}{\textbf{OOMD variance and squared bias} of mutual information estimators grouped by input dataset sizes $n$.}
        \label{tab:mi_grouped_by_row}
        \centering
        \resizebox{\linewidth}{!}{%
        \begin{tabular}{l|cc|cc}
        \toprule
          & \multicolumn{2}{c}{Bias$^2$} & \multicolumn{2}{c}{Variance} \\
         $n \in $ & [10, 150] & [150, 300] & [10, 150] & [150, 300] \\
        \midrule
        \midrule
        CCA & $.041$ {\scriptsize± $.007$} & $.037$ {\scriptsize± $.018$} & $.018$ {\scriptsize± $.003$} & $.006$ {\scriptsize± $.003$} \\
        KSG & $.044$ {\scriptsize± $.006$} & $.038$ {\scriptsize± $.001$} & $.014$ {\scriptsize± $.001$} & $.006$ {\scriptsize± $.000$} \\
        MINE & $1.601$ {\scriptsize± $.241$} & $.069$ {\scriptsize± $.008$} & $7.754$ {\scriptsize± $1.960$} & $.052$ {\scriptsize± $.027$} \\
        InfoNCE & $1.952$ {\scriptsize± $.664$} & $.064$ {\scriptsize± $.005$} & $9.513$ {\scriptsize± $7.801$} & $.004$ {\scriptsize± $.001$} \\
        NWJE & - & $.168$ {\scriptsize± $.031$} & - & $.074$ {\scriptsize± $.890$} \\
        
        \midrule
        ST$_{ABI}$ & $.027$ {\scriptsize± $.005$} & $.022$ {\scriptsize± $.007$} & $.029$ {\scriptsize± $.006$} & $.032$ {\scriptsize± $.005$} \\
        ST$_{NLL}$ & $.020$ {\scriptsize± $.006$} & $.003$ {\scriptsize± $.001$} & $.013$ {\scriptsize± $.002$} & $.011$ {\scriptsize± $.002$} \\
        ST$_{MSE}$ & $.015$ {\scriptsize± $.004$} & $.003$ {\scriptsize± $.000$} & $\mathbf{.010}$ {\scriptsize± $.001$} & $\mathbf{.001}$ {\scriptsize± $.000$} \\
        ST$_{MVUE}$ & $\mathbf{.012}$ {\scriptsize± $.003$} & $\mathbf{.002}$ {\scriptsize± $.000$} & $\mathbf{.010}$ {\scriptsize± $.001$} & $\mathbf{.001}$ {\scriptsize± $.000$} \\
        \bottomrule
        \end{tabular}
        }
    \end{minipage}
\end{figure}

\paragraph{Meta-Dataset Construction}
Following the method from \citep{NEURIPS2023_36b80eae}, we construct a meta-dataset using a two-stage process: (i) sampling a base distribution with known ground-truth MI, and (ii) applying MI-preserving transformations to induce variability. This procedure yields diverse distributions with known MI. To evaluate out-of-meta-distribution generalization (OOMD), we keep a set of base distributions and transformations excluded from training. In particular, we replicate the BMI benchmark \citep{NEURIPS2023_36b80eae} base distributions and transformations as the test set. The training meta-dataset has $150K$ pairs.\footnote{10 $\times$ less than the normality test experiment because it takes more time to generate, but was enough to yield strong results.}

\paragraph{Estimators Compared}
We benchmark variants of our approach against leading 1D MI estimators from \citep{NEURIPS2023_36b80eae}, including:
KSG \citep{PhysRevE.69.066138}, CCA \citep{pml2Book}, and three neural estimators: MINE \citep{pmlr-v80-belghazi18a}, InfoNCE \citep{repr_learning}, and NWJE \citep{NIPS2007_72da7fd6, NIPS2016_cedebb6e, pmlr-v97-poole19a}.
We train and evaluate four meta-statistical estimators sharing the same ST encoder architecture (270K parameters):
% \begin{itemize}
    \textbf{ST\textsubscript{MSE}}: trained with mean squared error loss;
    \textbf{ST\textsubscript{NLL}}: trained with Gaussian negative log-likelihood, predicting both mean and variance;
    \textbf{ST\textsubscript{MVUE}}: trained using the MVUE loss with trade-off \(\lambda = 0.5\), chosen on a small development set;
    \textbf{ST\textsubscript{ABI}}: combines the ST encoder with a coupling flow \citep{bayesflow_2020_original}, trained via the ABI generative loss using the BayesFlow library \citep{bayesflow_2023_software}.
% \end{itemize}

\paragraph{Evaluation Protocol}
We evaluate each estimator by measuring its \textbf{bias} (squared), \textbf{variance}, and \textbf{consistency} under controlled test conditions. Specifically, for each evaluation distribution \(\gamma\) selected in the test meta-dataset, we generate \(k = 100\) independent datasets of size \(n\) from $\gamma$. This setup allows us to estimate the estimator’s bias and variance for distribution $\gamma$. Then, the squared bias and variance are averaged over all test distributions in the meta-distribution. The results are reported in Table~\ref{tab:mi_grouped_by_row}. To evaluate generalization beyond the training regime, we further assess the consistency of our estimators by tracking their MSE as a function of input size \(n\) in \Figref{fig:mi_consistency}. The test meta-dataset contains $1.6$K distributions, each resampled $k=100$ times.

\paragraph{Results and Analysis}
In this OOMD setting, meta-statistical estimators consistently outperform baseline estimators, including extrapolation to sample sizes unseen during training. In particular, neural baselines (MINE, InfoNCE, and NJWE) perform poorly on small samples, whereas meta-statistical estimators remain robust. 
ST\textsubscript{MVUE} exhibits the lowest bias and generally achieves the best performance, demonstrating the efficacy of explicitly optimizing for minimum bias with the MVUE loss. This again shows that frequentist properties can be effectively controlled through loss function design.
ST\textsubscript{NLL} and ST\textsubscript{ABI} have the advantage of producing variance estimates, enabling uncertainty quantification. However, we observe that their confidence intervals are not often well calibrated OOMD (see details in Appendix~\ref{app:ci_cov}). Other methods (baselines or meta-statistical) would require external tooling to produce confidence intervals, such as conformal adaptation that uses a small sample of OOMD data to recalibrate the uncertainty estimates. 
% In line with the idea of our approach, amortized frequentist p--value estimation \citep{kadhim2023amortizedsimulationbasedfrequentistinference} could be used on top of learned estimators.

Our results highlight the promise of meta-statistical estimators. They generalize well beyond training conditions and can be trained efficiently (within one hour on a single GPU). Inference is orders of magnitude faster than existing neural estimators (ST\textsubscript{MVUE} offer inference speed-up of around $10^3 \times$ over MINE, InfoNCE, and NJWE and over $7 \times$ over ST\textsubscript{ABI}, see Appendix~\ref{app:mi_details}), opening the door to creating large-scale MI estimation for arbitrary input dimensions in the future.

%% file: 05_discussion.tex
Framing estimator design as a supervised learning problem over a meta-dataset offers a compelling link between machine learning and classical frequentist inference. Within this \emph{meta-statistical} perspective, familiar machine learning concepts acquire precise statistical meaning. For instance, minimizing the mean squared error (MSE) of an estimator $f$ for $\theta = g(\gamma)$ across a meta-distribution $\gamma \in \Gamma$ corresponds to minimizing the expected frequentist notion of bias and variance of $f$, averaged over $\Gamma$.
The notion of noise in regression, typically viewed as irreducible error, corresponds to the conditional variance ${\rm Var}_{\gamma}(\theta|X)$, the intrinsic uncertainty in inferring $\theta$ from data sampled from $\gamma$. This variance decreases with sample size $n$, governed by the Fisher information via the Cramér–Rao bound, making the meta-regression task inherently heteroscedastic: some distributions $\gamma$ and larger sample sizes are inherently easier.
By design, the permutation-invariant encoder $\phi$ must extract from $X$ a representation $\phi(X)$ that acts as a \textit{sufficient statistic} for $\theta$. 

\paragraph{Bayesian vs. Frequentist, when approximated via learning}
In frequentist inference, $\gamma$ is assumed fixed, and estimator properties are analyzed over resamples from $\gamma$. Extending to multiple $\gamma \in \Gamma$, a frequentist would seek guarantees that hold uniformly across $\Gamma$, without requiring a probabilistic structure on $\Gamma$. When approximated via learning (this work), we have to sample from $\Gamma$ but aim for generalization across $\gamma$'s (OOMD by varying $P_\Gamma$). 

By contrast, Bayesian inference treats $\theta$ as a random variable with a prior, and inference is performed by computing the Bayes rule given $X$. 
When approximated via learning (ABI), models are trained to amortize the computation of the Bayes rule by sampling $X$'s under a fixed simulator. This time, the generalization targets are unseen $X$’s drawn from the same simulator. The target is not generalization across simulators. 

Recent works proposed meta-ABI frameworks that learn to amortize Bayes rule computation from both $X$ and a prior $\pi(\theta)$ \citep{pmlr-v235-gloeckler24a, chang2025amortized}, aiming to learn functions $q(X, \pi(\theta))$ amortizing Bayes rule computation across priors and datasets $X$ within one simulator. Extending this further to generalize across simulators would imply learning a universal amortized Bayes rule, which remains a major challenge \citep{wu2020meta}. Meanwhile, the meta-statistical path allows learning dedicated models for each target property separately. We believe that both directions merit future exploration.

\paragraph{Limitations of this work}
Meta-statistical learning brings both the strengths and weaknesses of statistical learning. Notably, generalization out-of-meta-distribution (OOMD) is difficult to guarantee. However, it aligns with active research on out-of-distribution (OOD) generalization \citep{liu2023outofdistributiongeneralizationsurvey} and can directly import ideas from this field. Also, the meta-statistical learning problem provides a clean testbed for such OOD research, as it enables controlled variations in learning difficulty by tuning meta-prior shifts. 
% and choosing distributions with specific Fisher information.
% 
Interpretability remains an open issue: the learned inference algorithm becomes implicit and opaque \citep{molnar2022,electronics8080832,teney2022predicting}. Like LLMs, these estimators may exhibit unexpected failures and lack guarantees of statistical validity.
Finally, this work focuses on one-dimensional datasets. However, real-world inference often involves higher-dimensional data where classical estimators often fail. Scaling meta-statistical learning to these settings is a key next step, where meta-statistical learning is expected to bring large benefits over classical estimators. 

%% file: 11_acknowledgements.tex
\par This work was supported in part through the NYU IT High Performance Computing resources, services, and staff expertise. This material is based upon work supported by the National Science Foundation Graduate Research Fellowship under Grant No. DGE-2039655 (MR). Any opinion, findings, and conclusions or recommendations expressed in this material are those of the authors(s) and do not necessarily reflect the views of the National Science Foundation.
\par This work was supported by the Institute of Information \& Communications Technology Planning \& Evaluation (IITP) with a grant funded by the Ministry of Science and ICT (MSIT) of the Republic of Korea in connection with the Global AI Frontier Lab International Collaborative Research. This work was also supported by the Samsung Advanced Institute of Technology (under the project Next Generation Deep Learning: From Pattern Recognition to AI) and the National Science Foundation (under NSF Award 1922658).
\par This work was partially conducted within French research unit UMR 5217 and was supported by CNRS (grant ANR-22-CPJ2-0036-01) and by MIAI@Grenoble-Alpes (grant ANR-19-P3IA-0003).

%% file: 10_appendix.tex
\section{Decomposition of Meta-Statistical Regression Loss}
\label{app:math_regr}
Let $\theta \in \Theta \subset \mathbb{R}$ be a parameter of interest, defined as a function of an unobserved latent variable $\gamma \sim P(\gamma)$, i.e., $\theta = g(\gamma)$ for some measurable function $g$. Typically, $\gamma$ specifies one distribution among the set of possible distributions. Let $X = X_{1:n} = (X_1, \ldots, X_n) \sim P(X \mid \gamma)$ be a sample drawn i.i.d. given $\gamma$. The joint distribution over $(\gamma, X)$ is given by
\begin{equation}
P(\gamma, X) = P(\gamma) P(X \mid \gamma).
\end{equation}
This induces a joint distribution over $(\theta, X)$ via $\theta = g(\gamma)$.

Let $f(X)$ be an arbitrary estimator of $\theta$ based on the data $X$. Define the Bayes estimator (posterior mean) as
\begin{equation}
\phi(X) := \mathbb{E}[\theta \mid X],
\end{equation}
which minimizes the mean squared error (MSE) among all estimators with respect to $X$.

\subsection{MSE loss}
Now, we search for $f$ that minimizes the mean squared error of $f(X)$ under the joint distribution:
\begin{equation}
\text{MSE}(f) := \mathbb{E}_{\gamma, X}[(f(X) - \theta)^2].
\end{equation}

If $f$ is used as statistical estimator of $\theta$ when considering some distribution $\gamma$, its bias and variance (in the frequentist sense of resampling $X$ from one fixed distribution $\gamma$) are given by:
\begin{align*}
\mathrm{Bias}_{\gamma}(f) &= \mathbb{E}_{X|\gamma} [f(X) - g(\gamma)] = \mathbb{E}_{X|\gamma} [f(X)] - \theta \\
\mathrm{Var}_{\gamma}(f) &= \mathbb{E}_{X|\gamma} [(f(X) - \mathbb{E}_{X|\gamma} [f(X)])^2]
\end{align*}

\begin{proposition}[Conditional MSE decomposition into frequentists bias and variance]
The MSE of any estimator $f(X)$ admits the decomposition:
\begin{align*}
\mathbb{E}_{\gamma, X}[(f(X) - \theta)^2] &= \underbrace{\mathbb{E}_X[(f(X) - \phi(X))^2]}_{\text{approx. error to best estimator}} + \underbrace{\mathbb{E}_X[\mathrm{Var}(\theta \mid X)]}_{\text{irreducible posterior variance}} \\
&= \mathbb{E}_{\gamma} \underbrace{\mathrm{Var}_{\gamma}(f)}_{\text{frequentist variance of f}} + \mathbb{E}_{\gamma} [\underbrace{\mathrm{Bias}_{\gamma}(f)}_{\text{frequentist bias of f}}^2]
\end{align*}
\end{proposition}
It means two things: 
(i) when we optimize the MSE, we train to approximate the Bayes estimator, the irreducible noise is the fundamental uncertainty of $\theta$ from $X$, which decreases with $n$ but can be lower bounded by some degrees of non-identifiability of $\theta$
(ii) when we optimize the MSE, we optimize a function f to have low bias and variance in the frequentist sense when used as statistical estimator.

\begin{proof}
We begin by applying the law of total expectation:
\begin{equation}
\mathbb{E}_{\gamma, X}[(f(X) - \theta)^2] = \mathbb{E}_X\left[\mathbb{E}_{\theta \mid X}[(f(X) - \theta)^2]\right].
\end{equation}

Inside the expectation, write:
\begin{equation}
(f(X) - \theta)^2 = (f(X) - \phi(X) + \phi(X) - \theta)^2.
\end{equation}
Expanding this yields:
\begin{equation}
(f(X) - \theta)^2 = (f(X) - \phi(X))^2 + (\phi(X) - \theta)^2 + 2(f(X) - \phi(X))(\phi(X) - \theta).
\end{equation}
Taking the conditional expectation given $X$:
\begin{equation}
\mathbb{E}_{\theta \mid X}[(f(X) - \theta)^2] = (f(X) - \phi(X))^2 + \mathbb{E}_{\theta \mid X}[(\phi(X) - \theta)^2] + 2(f(X) - \phi(X)) \underbrace{\mathbb{E}_{\theta \mid X}[\phi(X) - \theta]}_{= 0}.
\end{equation}
Hence:
\begin{equation}
\mathbb{E}_{\theta \mid X}[(f(X) - \theta)^2] = (f(X) - \phi(X))^2 + \mathrm{Var}(\theta \mid X),
\end{equation}
and thus:
\begin{equation}
\mathbb{E}_{\gamma, X}[(f(X) - \theta)^2] = \mathbb{E}_X[(f(X) - \phi(X))^2] + \mathbb{E}_X[\mathrm{Var}(\theta \mid X)] \\
\end{equation}

This completes the first decomposition.

The second decomposition is the standard conditional bias variance decomposition. 
Start by writing:
\begin{equation}
\mathbb{E}_{\gamma, X}[(f(X) - \theta)^2] = \mathbb{E}_\gamma\left[ \mathbb{E}_{X \mid \gamma}[(f(X) - \theta)^2] \right].
\end{equation}

Inside the inner expectation, define $\mu_f(\gamma) := \mathbb{E}_{X \mid \gamma}[f(X)]$, so:
\begin{equation}
(f(X) - \theta)^2 = (f(X) - \mu_f(\gamma) + \mu_f(\gamma) - \theta)^2.
\end{equation}
Expanding:
\begin{equation}
(f(X) - \theta)^2 = (f(X) - \mu_f(\gamma))^2 + (\mu_f(\gamma) - \theta)^2 + 2(f(X) - \mu_f(\gamma))(\mu_f(\gamma) - \theta).
\end{equation}
Now take expectation over $X \mid \gamma$:
\begin{align*}
\mathbb{E}_{X \mid \gamma}[(f(X) - \theta)^2] 
&= \mathbb{E}_{X \mid \gamma}[(f(X) - \mu_f(\gamma))^2] + (\mu_f(\gamma) - \theta)^2 \\
&\quad + 2(\mu_f(\gamma) - \theta) \mathbb{E}_{X \mid \gamma}[f(X) - \mu_f(\gamma)].
\end{align*}
Note that:
\begin{equation}
\mathbb{E}_{X \mid \gamma}[f(X) - \mu_f(\gamma)] = \mathbb{E}_{X \mid \gamma}[f(X)] - \mu_f(\gamma) = 0,
\end{equation}
so the cross term vanishes $2(\mu_f(\gamma) - \theta) \cdot 0 = 0$
Thus:
\begin{equation}
\mathbb{E}_{X \mid \gamma}[(f(X) - \theta)^2] = \mathrm{Var}_{\gamma}(f) + \mathrm{Bias}_{\gamma}(f)^2.
\end{equation}
Now average over $\gamma$:
\begin{equation}
\mathbb{E}_{\gamma, X}[(f(X) - \theta)^2] = \mathbb{E}_\gamma[\mathrm{Var}_{\gamma}(f)] + \mathbb{E}_\gamma[\mathrm{Bias}_{\gamma}(f)^2].
\end{equation}
\end{proof}

\section{Architecture Comparison with Descriptive Tasks}
\label{app:desc_details}

\subsection{Data Generation}
In descriptive tasks, the label \( y \) of a dataset \( \mathcal{D} \) is the output of an algorithm \( A \) applied to \( \mathcal{D} \), i.e., \( y = A(\mathcal{D}) \). Simple tasks like median or correlation serve as unit testing of meta-statistical models, since there is no need for a meta-statistical model to compute such algorithms. However, for more computationally intensive algorithms, such as optimal transport, meta-statistical models could serve as fast approximations~\cite{amos2023tutorialamortizedoptimization}. For datasets \( \mathcal{D} \in \mathbb{R}^{n \times m} \), we consider four descriptive tasks: the \textbf{per-column median} label \( y \in \mathbb{R}^m \) consists of the medians of each column. The \textbf{Pearson correlation} coefficient \( y \in \mathbb{R} \) is computed between the two columns. The \textbf{win rate} (Bradley-Terry) is the fraction of rows where the value in the first column exceeds that in the second: \( y = \frac{1}{n} \sum_{i=1}^n \mathbb{I}(\mathcal{D}_{i,1} > \mathcal{D}_{i,2})\), where \( \mathbb{I}(\cdot) \) is the indicator function. Finally, the 1D \textbf{optimal transport} (OT) label \( y \in \mathbb{R} \) is the optimal transport cost between the empirical distributions of the two columns.

\paragraph{In-Meta-Distribution}
The set of distributions used to generate datasets during training is parameterized as follows:
\begin{itemize}
    \item \texttt{normal}:: It has two parameters: the mean and the variance. Mean values are sampled from $\texttt{[-3, 3]}$, and variances are sampled from $\texttt{[0.1, 1.5]}$. 
    \item \texttt{uniform}:: It has two parameters: the lower bound and the upper bound. The lower bounds are sampled from $\texttt{[-3.5, -0.5]}$ and the upper bounds from $\texttt{[0.5, 3.5]}$.
    \item \texttt{beta}:: It has two parameters: $a$ and $b$. Parameters $a$ and $b$ are sampled from $\texttt{[1, 3]}$ and $\texttt{[2, 5]}$, respectively.
    \item \texttt{exponential}:: It has one parameter: scale sampled from $\texttt{[1, 2]}$.
\end{itemize}

\paragraph{Out-of-Meta-Distribution}
The set of distribution used to test models for unseen distribution families is parametrized as follows:
\begin{itemize}
    \item \texttt{gamma}:: It has two parameters: shape and scale. Shape parameters are sampled from $\texttt{[1, 5]}$, and scale parameters from $\texttt{[1, 2]}$.
    \item \texttt{log-normal}:: It has two parameters: mean and variance. Means are sampled from $\texttt{[0, 1]}$, and standard deviations from $\texttt{[0.5, 0.75]}$.
\end{itemize}

\paragraph{Meta-dataset properties}
Once a distribution $P_X$ has been sampled, we use it to sample one dataset. In general, we could sample several dataset per distributions but we prefer to sample only one to maximize the diversity of distributions seen during training. Each dataset is defined by the number of rows (\texttt{n\_row\_range}), sampled uniformly from the range \texttt{[5, 300]}. During testing, we explore longer lengths to test the generalization of meta-statistical models. This results in a meta-datapoint. We them sample many meta-datapoints to build a meta-datasets with the following split sizes: 30K training, 300 validation, 3K for testing in-meta-distribution and 3K for testing out-of-meta-distribution.
 
\subsection{Architectures}
We compare several encoder architectures. As a baseline, we include an LSTM \citep{hochreiter1997long}, which treats datasets as sequences and thus lacks permutation invariance. The Vanilla Transformer (VT), in contrast, is permutation-invariant due to the symmetry of the attention mechanism. The Set Transformer \citep{lee2019set} reduces the quadratic cost of attention by introducing a fixed number of learned inducing points on which attention is performed. We adopt an improved version, Set Transformer 2 (that we call ST here) \citep{zhang2022set}, which uses SetNorm \citep{ba2016layer} to preserve permutation invariance while improving empirical training convergence. We compare two variants of ST2: with 16 or 32 inducing points. ST2(16) is the fastest model for both training and inference. In Appendix~\ref{app:eff}, we show that VT scales quadratically, while LSTM and ST2 scale linearly. For consistency in reporting, we compare models with approximately the same number of parameters (\( \sim 10K \) in this section).
As we focused on tiny architectures, each model can be trained on GPU in less than an hour. Models are trained with Adam and learning rate of $1e-5$ and weight decay of $1e-6$.

\subsection{Results}
\paragraph{In-meta-distribution performance}  
Table~\ref{tab:performance_comparison} shows the MSE of the four meta-statistical models on a test set sampled from the same meta-distribution as the training data. All models approximate the descriptive tasks well, but the LSTM-based model, lacking permutation invariance, performs worse than attention-based models. 
Notably, ST2, despite being much faster than VT, narrowly outperforms it. Given its strength and efficiency, ST2 is our main model in the rest of the paper, with VT considered as an alternative baseline.

\begin{figure}[t] % 't' places the figure at the top of the page
    \centering
    % Subfigure (a)
    \begin{subfigure}[t]{0.49\textwidth} % Adjust width as needed
        \centering
        \includegraphics[width=\textwidth]{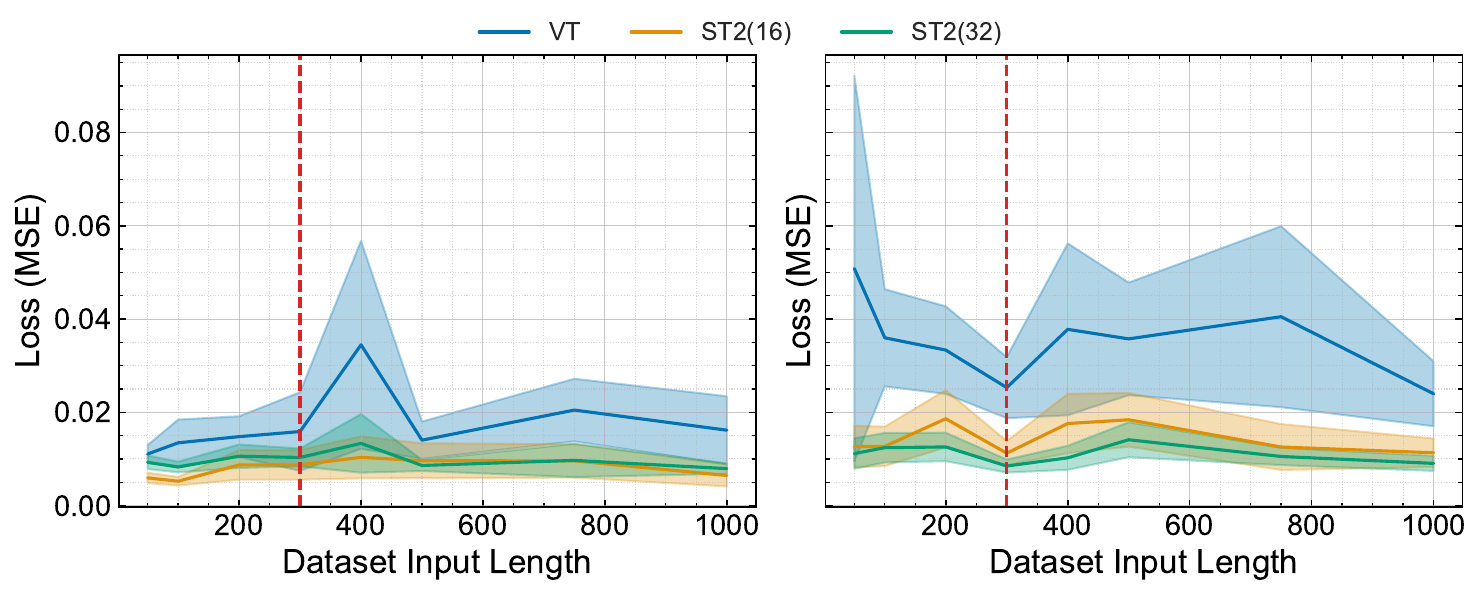}
        \caption{\texttt{Win-rate} prediction}
        \label{fig:subfig_a_app}
    \end{subfigure}
    \hfill
    % Subfigure (b)
    \begin{subfigure}[t]{0.49\textwidth}
        \centering
        \includegraphics[width=\textwidth]{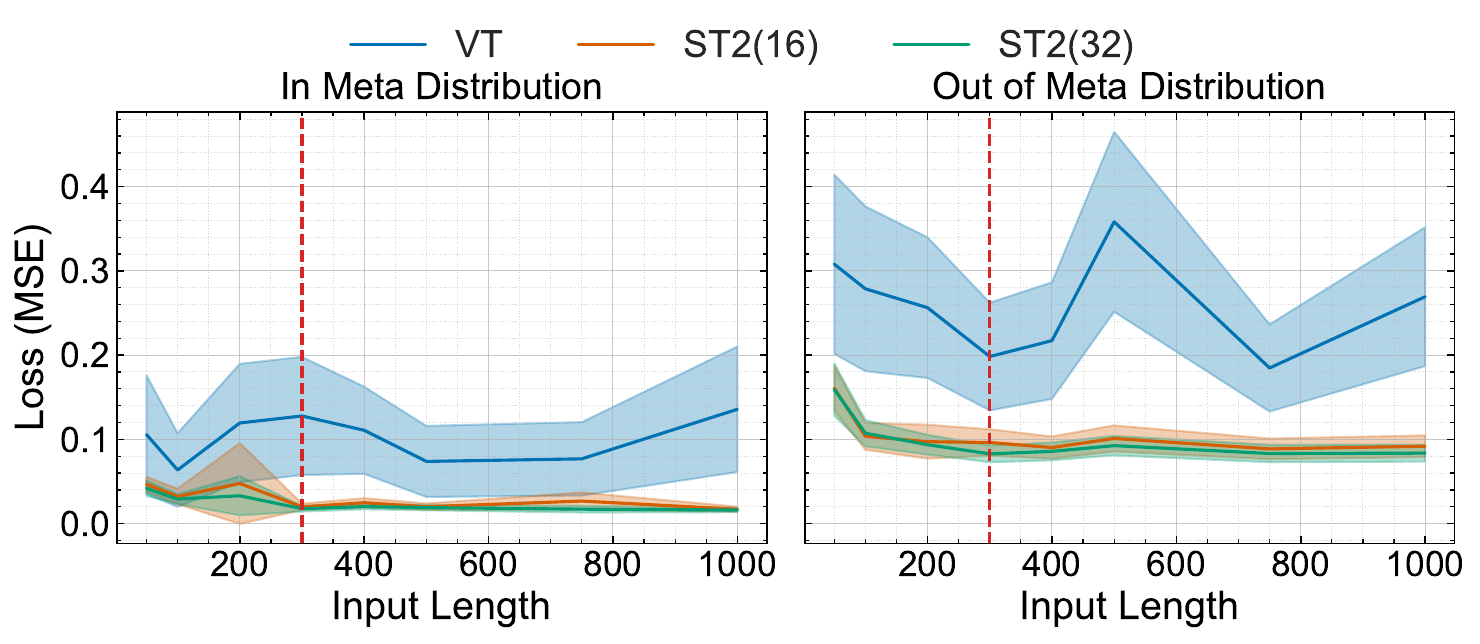} 
        \caption{\texttt{Median} prediction}
        \label{fig:subfig_b_app}
    \end{subfigure}
    \vspace{0.5cm} % Add vertical spacing between rows of subfigures
    % Subfigure (c)
    \begin{subfigure}[t]{0.49\textwidth}
        \centering
        \includegraphics[width=\textwidth]{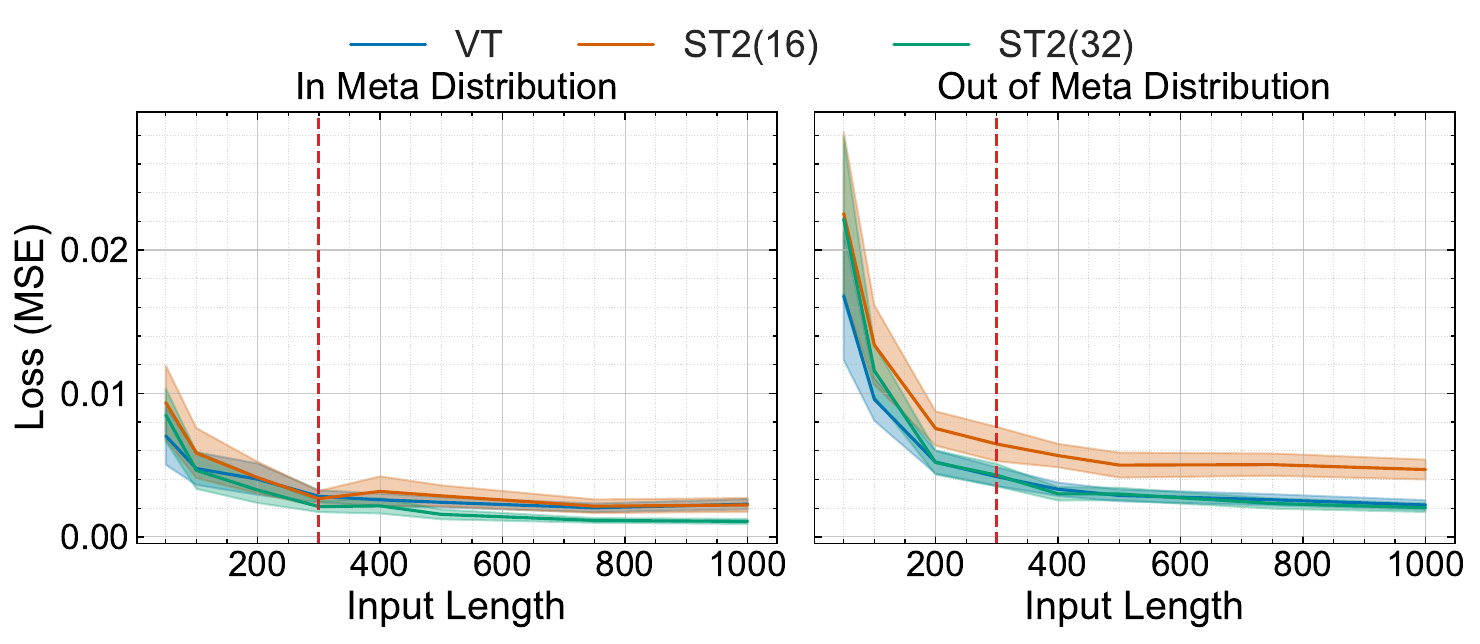}
        \caption{\texttt{Correlation} prediction}
        \label{fig:subfig_c_app}
    \end{subfigure}
    \hfill
    % Subfigure (d)
    \begin{subfigure}[t]{0.49\textwidth}
        \centering
        \includegraphics[width=\textwidth]{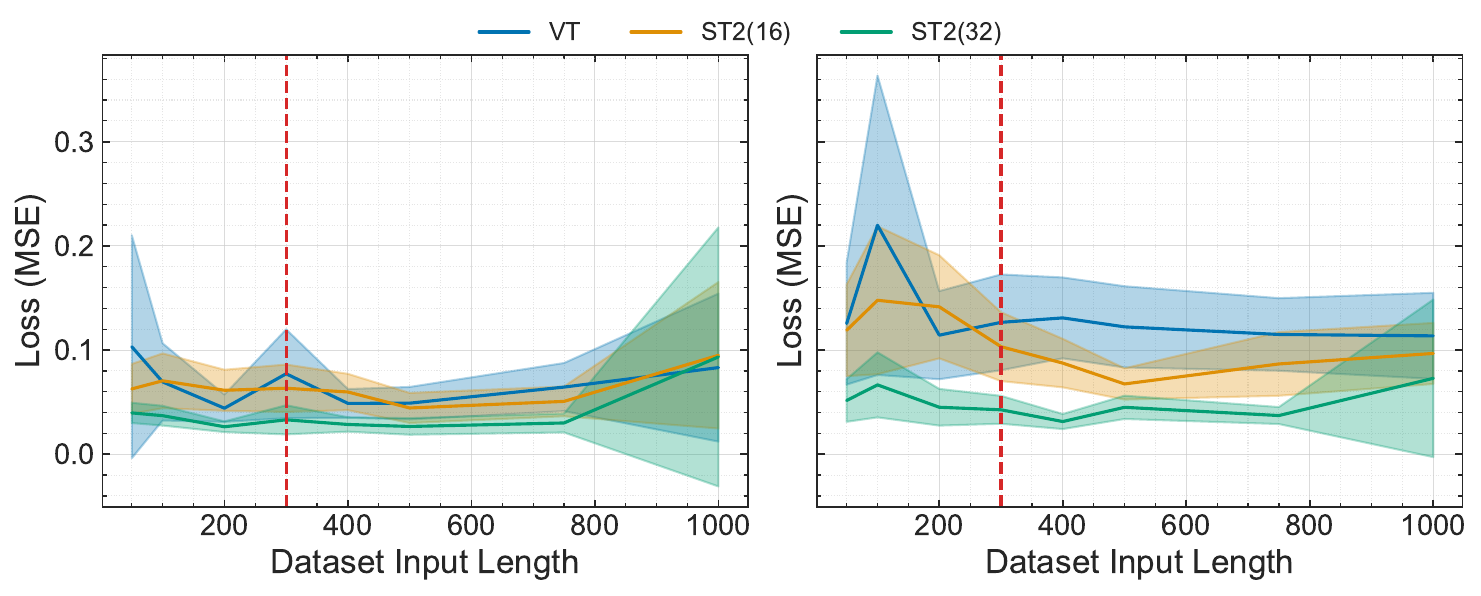} 
        \caption{\texttt{Optimal Transport} prediction}
        \label{fig:subfig_d_app}
    \end{subfigure}

    % \vspace{-0.5cm}
    % Main caption for the figure
    \caption{\textbf{Generalization Across Dataset Lengths and Meta-Distributions.} For each subplot, the left panel illustrates the performance of meta-statistical models on test datasets that vary in input length, including lengths not observed during training, while remaining within the training meta-distribution. For each subplot, the right panel presents the same comparison but for test datasets sampled from entirely new meta-distributions, with distributions unseen during training. Note that LSTM is excluded because its errors are an order of magnitude higher.}
    \label{fig:generalization}
\end{figure}

\paragraph{Generalization Performance}  
We evaluate meta-statistical models' generalization capabilities on two aspects:  (i) \textbf{Out-of-Meta-Distribution (OoMD)}: Datasets from unseen distributions. (ii) \textbf{Length Generalization}: Datasets with lengths outside the training range. \Figref{fig:generalization} shows strong length generalization, where models maintain their performance for larger datasets than seen during training, both IMD and OoMD. 
They are also robust to OoMD datasets despite a small performance degradation. Manual inspection reveals that the degradation mainly comes from cases where the magnitude of the input values exceeds the range seen during training.

\begin{table}[t]
\centering
% \resizebox{0.95\columnwidth}{!}{
\begin{tabular}{@{}l|c|c|c|c@{}}
\toprule
& \textbf{Median} & \textbf{Corr} & \textbf{WinRate (BT)} & \textbf{OT (1D)} \\ 
\midrule
\midrule
LSTM
& $2.9e^{-1}$ \scriptsize{$\pm 0.8$}
& $5.9e^{-2}$ \scriptsize{$\pm 1.5$}
& $4.4e^{-2}$ \scriptsize{$\pm 0.9$}
& $8.5e^{-2}$ \scriptsize{$\pm 2.9$}\\
VT
& $\mathbf{6.0e^{-2}}$ \scriptsize{$\pm 1.9$}
& $\mathbf{9.2e^{-3}}$ \scriptsize{$\pm 4.6$}
& $7.1e^{-3}$ \scriptsize{$\pm 1.5$}
& $\mathbf{5.5e^{-2}}$ \scriptsize{$\pm 1.4$}\\
ST2(16)
& $\mathbf{4.2e^{-2}}$ \scriptsize{$\pm 1.7$}
& $\mathbf{7.5e^{-3}}$ \scriptsize{$\pm 2.8$}
& $\mathbf{2.9e^{-3}}$ \scriptsize{$\pm 1.2$}
& $\mathbf{4.5e^{-2}}$ \scriptsize{$\pm 1.9$}\\
ST2(32)
& $\mathbf{4.4e^{-2}}$ \scriptsize{$\pm 0.9$}
& $\mathbf{9.1e^{-3}}$ \scriptsize{$\pm 5.1$}
& $\mathbf{1.6e^{-2}}$ \scriptsize{$\pm 0.5$}
& $\mathbf{3.0e^{-2}}$ \scriptsize{$\pm 1.5$}\\
\bottomrule

\end{tabular}
% }
\caption{Performance comparison meta-statistical models across tasks, measured by Mean Squared Error with respect to correct output on the test set. \textbf{Bold} indicates no significant difference with the best model.}
\label{tab:performance_comparison}
\end{table}

\subsection{Examples of Training Curves}
\label{app:training_curves}
\begin{figure}[t]
    \centering
    \includegraphics[width=0.6\columnwidth]{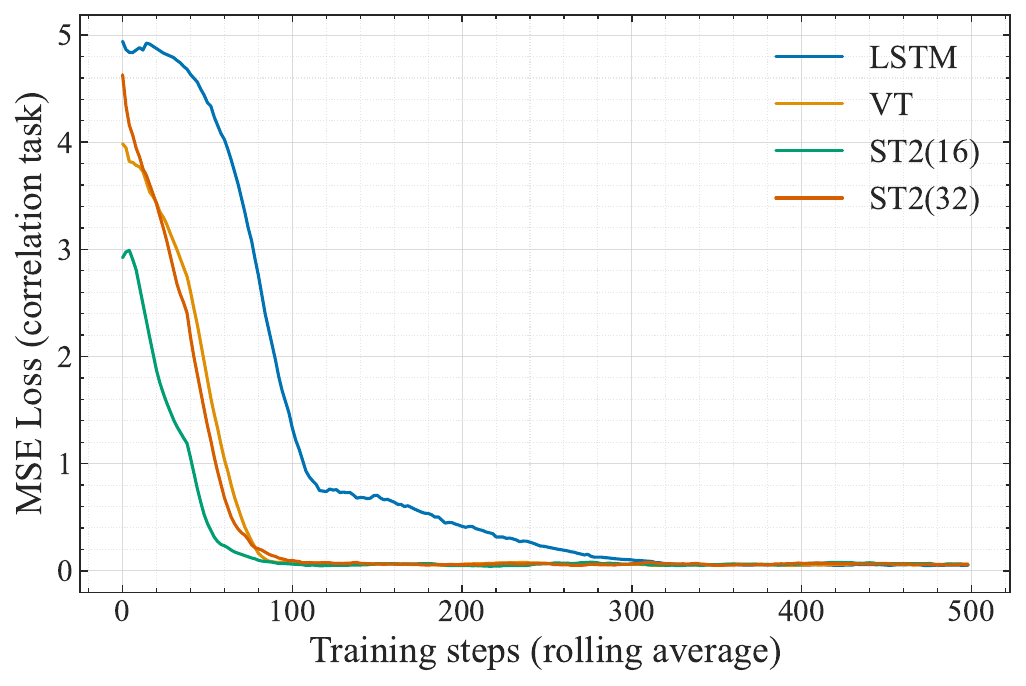} % Replace with your image file
    \caption{\textbf{Training curves:} Comparison of training convergence of meta-statistical models on the correlation task.}
    \label{fig:training_curves}
\end{figure}

For meta-statistical models of approximately the same size ($\approx$ 10K parameters), we compare their convergence during training on the task of predicting the correlation between variable A and variable B, the two columns of the dataset. We report the results in \Figref{fig:training_curves}.

\subsection{Details about Efficiency}
\label{app:eff}
\begin{figure}
    \centering
    \includegraphics[width=0.6\columnwidth]{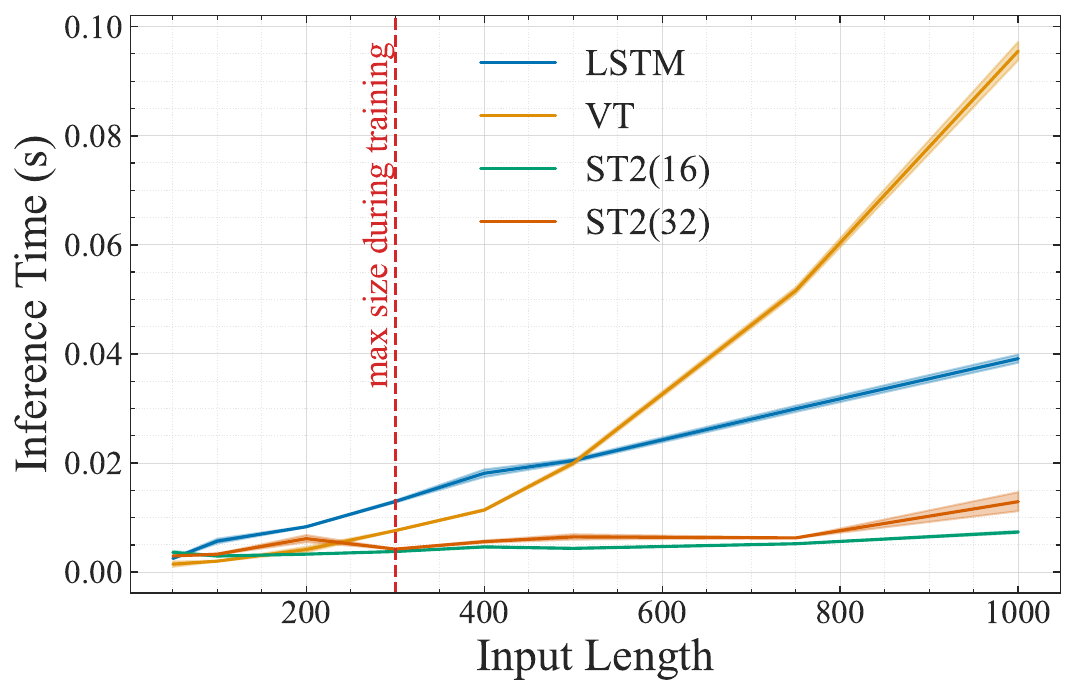}
    \caption{\textbf{Inference time} comparison of meta-statistical models per batch as a function of input dataset length. Models have similar parameter counts $\approx 10K$).}
    \label{fig:inference_time}
\end{figure}

In \Figref{fig:inference_time}, we present the inference time of meta-statistical models as a function of the input dataset size \( n \). As expected, the VT scales quadratically, whereas LSTM and ST2 variants scale linearly with slopes in favor of ST2.
We also compare the efficiency per parameter. For this we compute both the training and inference time of each model per batch averaged over 1K batches, and normalized by the number of parameters in the model. The results are reported in Table~\ref{tab:times}. Given the strong performance of ST2 and the clear computational advantage we see it as strong meta-statistical architecture.

\begin{table}[t]
\centering
 \begin{tabular}{@{}l|c|c@{}}
    \toprule
     & Training Time & Inference Time \\
    \midrule
    \midrule
    VT & $2.2e^{-5} $ \scriptsize{$\pm 1.1$} & $2.7e^{-3}$ \scriptsize{$\pm 1.2$} \\
    LSTM & $6.5e^{-6} $ \scriptsize{$\pm 1.0$} & $8.5e^{-4}$ \scriptsize{$\pm 3.9$} \\
    ST2(32) & $5.9e^{-6} $ \scriptsize{$\pm 6.9$} & $1.5e^{-3}$ \scriptsize{$\pm 2.4$} \\
    ST2(16) & $1.7e^{-6} $ \scriptsize{$\pm 0.2$} & $2.2e^{-4}$ \scriptsize{$\pm 0.9$} \\
    \bottomrule
\end{tabular}
\caption{Comparison of training and inference times for various models, normalized per batch per number of parameters.}
\label{tab:times}
\end{table}

\section{Details about Normality Tests Experiments}
\label{app:norm_test}

\subsection{Meta-Dataset Creation}
We construct a meta-dataset by generating datasets labeled with the ground truth binary indicator of normality, using a diverse set of alternative distributions. In previous studies, the uniform distribution was used the contrast distribution \citep{razali2011power}. To create a meta-dataset, we flip an unbiased coin to decide if we draw from the normal or not. If the outcome is normal, we sample the mean and variance of the Gaussian distribution; otherwise, we sample uniformly at random one distribution from the set of non-Gaussian distributions and then draw its parameters from the predefined prior. A dataset size \( n \) is then drawn uniformly at random from the range \( [5, 150] \), and the dataset \( X \) is sampled with \( n \) rows. During testing, the dataset sizes range up to $300$ and the testing involves resampling $k=100$ times from the same distribution to compute quantities of interest such as power and calibration error.

For all estimators, the datasets $X$ are standardized to ensure empirical mean of zero and empirical variance of one. It is common practice for classical estimators and prevents the meta-statistical estimators to overfit potential spurious correlations between value ranges and classification answer.

The set of distributions used to generate datasets during training and testing is parameterized as follows:
\begin{itemize}
    \item \texttt{Gaussian}:: It has two parameters: the mean and the standard deviation. Means are sampled from $\texttt{[-3, 3]}$, and standard deviations from $\texttt{[0.1, 3]}$.
    \item \texttt{Uniform}:: It has two parameters: the lower and upper bounds. Lower bounds are sampled from $\texttt{[-2, 0]}$ and upper bounds from $\texttt{[0, 2]}$.
    \item \texttt{Exponential}:: It has one parameter: the scale, sampled from $\texttt{[0.5, 2]}$.
    \item \texttt{Beta}:: It has two parameters: $a$ and $b$. Parameter $a$ is sampled from $\texttt{[0.5, 5]}$ and parameter $b$ from $\texttt{[0.5, 5]}$.
    \item \texttt{Log-Normal}:: It has two parameters: the mean and the standard deviation of the underlying normal distribution. Means are sampled from $\texttt{[-1, 1]}$ and standard deviations from $\texttt{[0.5, 1.5]}$.
    \item \texttt{Gamma}:: It has two parameters: the shape and the scale. Shape parameters are sampled from $\texttt{[1, 5]}$ and scales from $\texttt{[0.5, 2]}$.
    \item \texttt{Triangular}:: It has three parameters: the left bound, the mode, and the right bound. Left bounds are sampled from $\texttt{[-1, 0]}$, modes from $\texttt{[0.1, 1]}$, and right bounds from $\texttt{[1, 2]}$.
    \item \texttt{Cauchy}:: It has two parameters: the location $x_0$ and the scale $\gamma$. Locations are sampled from $\texttt{[0, 1]}$, and scales from $\texttt{[0.5, 2]}$.
\end{itemize}
The uniform, log-normal and Triangular are seen only during testing (OoMD) while the other families are seen during training (IMD).

\subsection{Details about the Training of Meta-Statistical Models}
The experiments compared the performance of meta-statistical models where the encoder is an ST2 architecture with 4 input layers of 24 heads with 32 inducing points. The dimensionality of MLPs is 72. The models are trained with lr$=3e-4$, weight decay $1e-5$ with gradient clipping at $3.5$ on a meta-dataset of 1.5M examples for two epochs. Each model is trained in less than 2 hours on a single GPU.

\begin{figure}
    \centering
    \includegraphics[width=\linewidth]{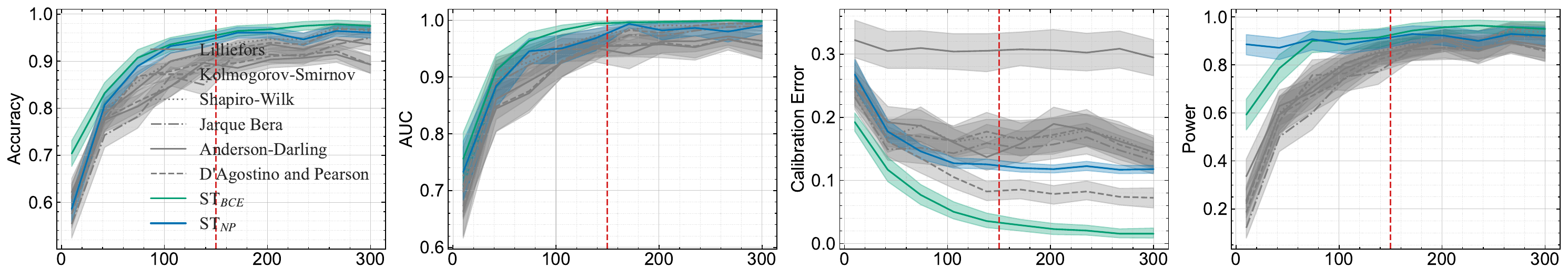}
    \caption{\textbf{OOMD evaluation} of all normality tests as a function of the input dataset size. (Red line: training cut-off)}
    \label{fig:norm_results_app}
\end{figure}

\subsection{Full results}
In figure~\ref{fig:norm_results_app}, we report the same plot as the main paper, but with all the baselines. The conclusions remain the same.

\subsection{Performance per distribution families}
\begin{figure}
    \centering
    \includegraphics[width=0.7\linewidth]{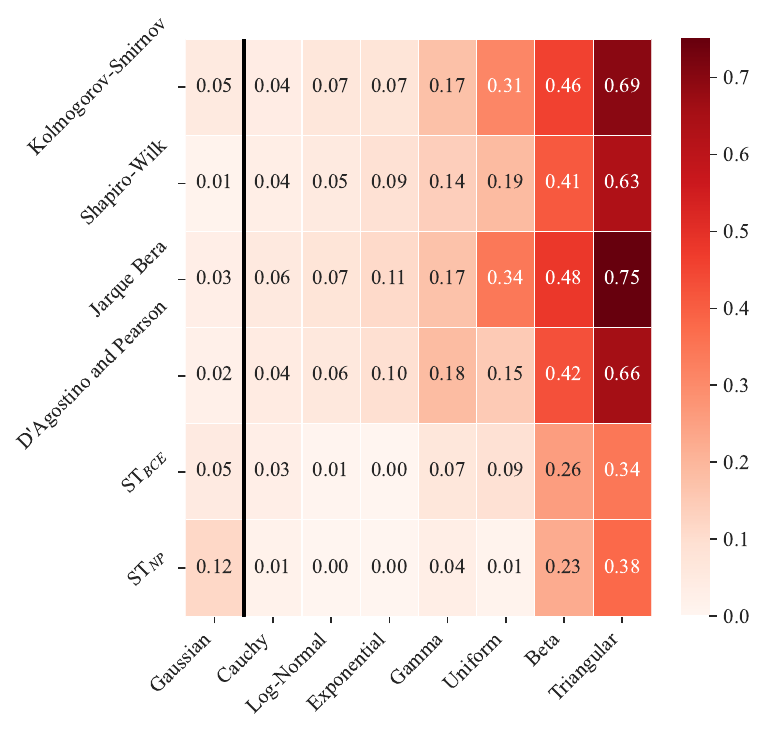}
    \caption{Classification error rates per distribution families.}
    \label{fig:error_rates}
\end{figure}

We report the classification error rate per distribution families for all estimators in Figure~\ref{fig:error_rates}. Interestingly, we see the different behavior favored by the two different losses $\mathcal{L}_{BCE}$ and $\mathcal{L}_{NP}$.

\section{Details about Mutual Information Experiments}
\label{app:mi_details}

\subsection{Details of Meta-Dataset Creation}
We construct a meta-dataset inspired by the benchmark methodology in \cite{NEURIPS2023_36b80eae}, where distributions with ground-truth MI are generated in two steps: (i) by sampling a distribution with known MI, (ii) potentially applying MI-preserving transformation. This process creates complex distributions and datasets with known MI. WE use different base-distribution and MI-preserving transformation between in-meta-distribution and out-of-meta-distribution. 

Base distributions are: \texttt{binormal}, \texttt{bimodal\_gaussians}, \texttt{binormal}, \texttt{bistudent}, and \texttt{additive\_noise}
The MI-preserving transformations are: \texttt{[base, wigglify, halfcube, asinh, normal\_cdfise]}. 
For the test set, we re-use the 1D BMI benchmark~\citep{NEURIPS2023_36b80eae}. For sampling both the training and testing meta-datasets, we use the tool they provided: \url{https://github.com/cbg-ethz/bmi}.

\subsection{Details about the Training of Meta-Statistical Models}
The experiments compared the performance of meta-statistical models where the encoder is SetTransformer 2 architecture with 3 input layers of 16 heads with 32 inducing points. The dimensionality of MLPs is 64. The models are trained with lr$=1e-4$, weight decay $1e-5$ with gradient clipping at $3.5$ on a meta-dataset of 1.5M examples for two epochs. Each model is trained in less than 2 hours on a single GPU.

\subsection{Additional Results}
\label{app:ci_cov}
\begin{figure}
    \centering
    \includegraphics[width=\linewidth]{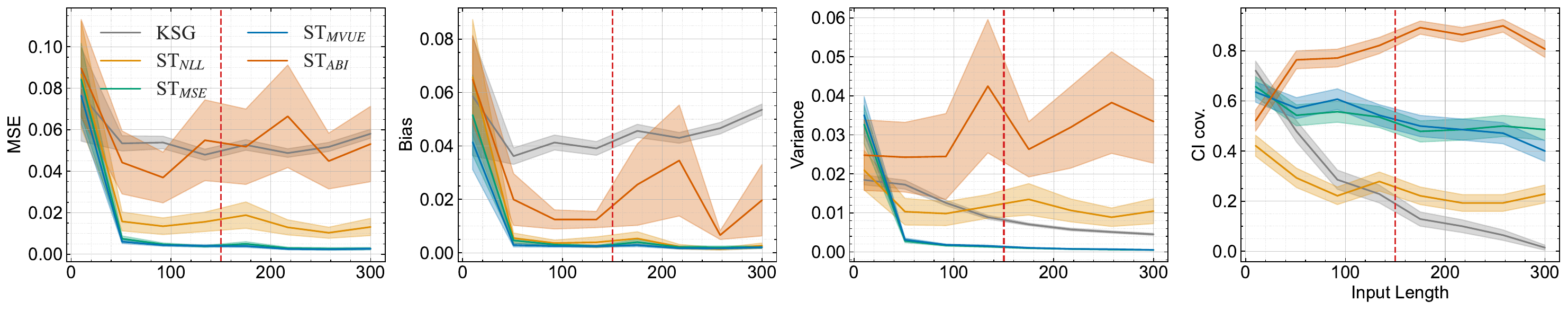}
    \caption{Consistency of MI estimators for MSE, bias, variance, and CI coverage.}
    \label{fig:app_mi_const}
\end{figure}

We report as a function of sample size $n$, not only the MSE but also the bias (squared), the variance, and the coverage of the confidence intervals. To compute confidence intervals, we can use bootstrap resampling (as done to compute bias and variance) and estimate the confidence interval at a given threshold $\alpha$ (that we fix to 0.05).
For ST\textsubscript{NLL} and ST\textsubscript{ABI}, we can obtain the confidence interval directly from their parametrized distribution over $\theta$.
The confidence interval can either cover or not the true MI of this distribution. This is reported in \Figref{fig:app_mi_const}.

All estimators seem consistent for MSE, bias, and variance. However, confidence intervals are poorly calibrated across (which also concerns the baselines). In particular, even generative modeling approaches do not represent well the uncertainty of $\theta|X$ in the OOMD setting.
This is a standard issue in machine learning that can be approach by additional recalibration such as conformal prediction \cite{angelopoulos2022gentleintroductionconformalprediction}.

\begin{figure}
    \centering
    \includegraphics[width=\linewidth]{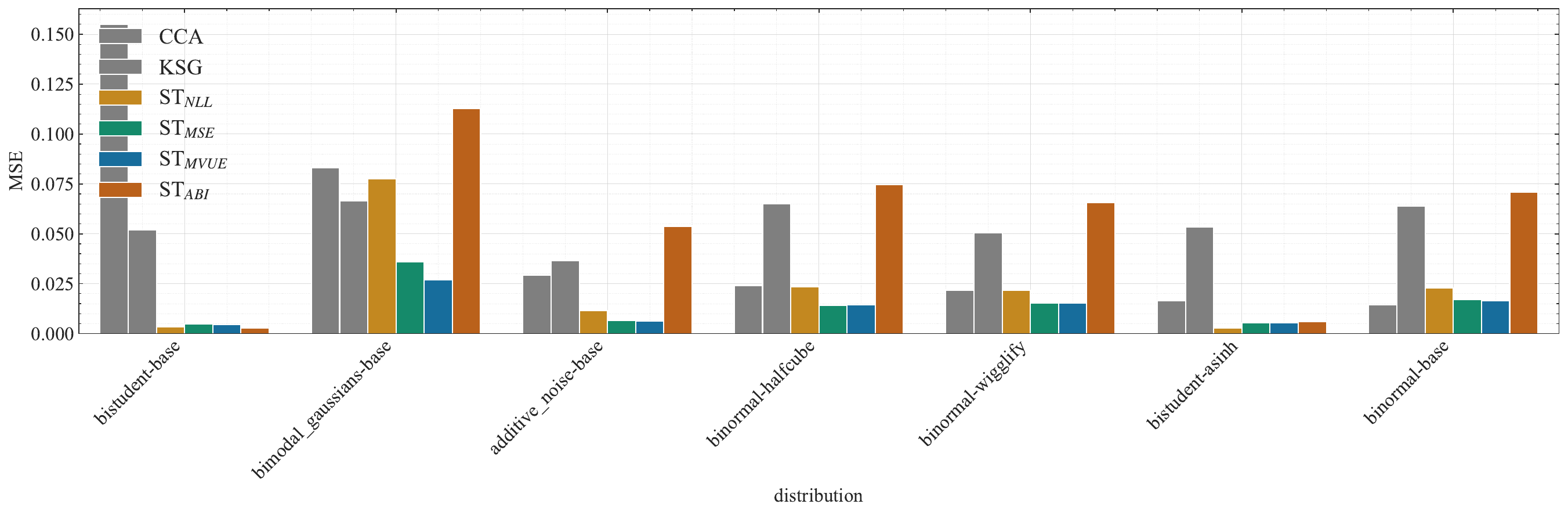}
    \caption{MSE scores of MI estimators across distribution families in the test set.}
    \label{fig:mi_per_distrib}
\end{figure}

Also, we report the MSE performance of estimators across distribution families within the test set in \Figref{fig:mi_per_distrib} 

\begin{wraptable}{r}{0.3\linewidth}
\centering
\vspace{-1em}
\begin{tabularx}{\linewidth}{l l}
\toprule
 & \textbf{Speedup} \\
\midrule
ST\textsubscript{ABI} & $\times$ 7.4 \\
MINE & $\times$ 1008.4 \\
InfoNCE & $\times$ 870.2 \\
NJWE & $\times$ 1180.6 \\
\bottomrule
\end{tabularx}
% \vspace{-1em}
\caption{Inference speed of ST\textsubscript{MVUE} compared to other neural methods.}
\label{tab:inference-time}
\end{wraptable}

\subsection{Inference speed-up}
Meta-statistical estimators offer a significant speed up over existing neural estimators since they can amortize the inference cost as a single forward pass. Furthermore, purely discriminative approaches like ST\textsubscript{MVUE} oST\textsubscript{MSE} also offer a significant speed-up over amortized but generative modeling of ST\textsubscript{ABI}. This results are documented in Table~\ref{tab:inference-time}